\documentclass[11pt,a4paper]{ubtreport}
\usepackage{graphicx}
\usepackage{tcolorbox}
\tcbuselibrary{skins}
\usepackage{xcolor}
\usepackage{ulem}
\usepackage{booktabs}
\usepackage{multirow}
\usepackage{graphicx}
\usepackage{tikz}
\usepackage{subcaption}
\usetikzlibrary{
    arrows.meta,
    positioning,
    fit,
    calc,
    backgrounds
}
\companyname{Thinker Team, Ubtech}
\doctype{Technical Report}
\reptitle{
Thinker-WM: A Unified World Model with Shared Multimodal Backbone for Embodied Intelligence
}
\title{}
\productname{Thinker-WM}

\keyword{World Model}
\keyword{Video-Action Model}
\keyword{Flow Matching}
\keyword{Shared Multimodal Representation}
\keyword{Robot Data Pipeline}
\keyword{Multi-Node Training}

\begin{document}

\maketitle


\begin{abstract}

\noindent World models have emerged as a promising paradigm for embodied AI, yet existing systems often remain fragmented in the modalities they model and the functions they support. In particular, visual dynamics, geometric structure, and executable action are commonly coupled through separate interfaces or downstream modules rather than being jointly represented and evolved within a shared dynamics model.

In this work, we present \textbf{Thinker-WM}, a unified multimodal world model that jointly models and generates RGB, metric depth, and robot action through a shared intermediate \textbf{Unified Dynamics Transformer} with modality-specific encoders and decoders. The model performs multimodal unification at the representation level: appearance, geometry, and action are transformed into a common dynamic representation and jointly evolved by shared parameters, while their distinct output structures are preserved. The same parameterization supports multiple functions, including joint video–action generation, world prediction, policy inference, action-conditioned generation, video-to-action generation, and inverse dynamics. To examine whether the shared representation captures action-related information, we introduce a lightweight motion probe that shows the frozen visual representation contains recoverable motion information.

Thinker-WM is trained on a heterogeneous corpus comprising approximately \textbf{20K} hours of egocentric video, \textbf{10K} hours of real-robot data, and \textbf{1K} hours of simulation data, including approximately \textbf{4K} hours of real-robot data with aligned depth supervision.  Extensive experiments demonstrate strong performance across embodied world-model evaluations. Thinker-WM ranks \textbf{first} on \textbf{EWMBench}, \textbf{WorldOlympiad} and \textbf{WorldArena}, while controlled studies further demonstrate the benefits of action conditioning, geometric interaction, and multi-function training within the shared representation. 

Code is available at \href{https://github.com/UBTECH-Robot/Thinker-WM}{\textit{\textbf{GitHub}}} and \href{https://thinkercosmos.ubtrobot.com/thinker-wm/}{\textit{\textbf{Project Page}}}.

\end{abstract}


\section{Introduction}
\label{sec:intro}

\begin{figure}[htbp]
    \centering
    \includegraphics[width=1\textwidth]{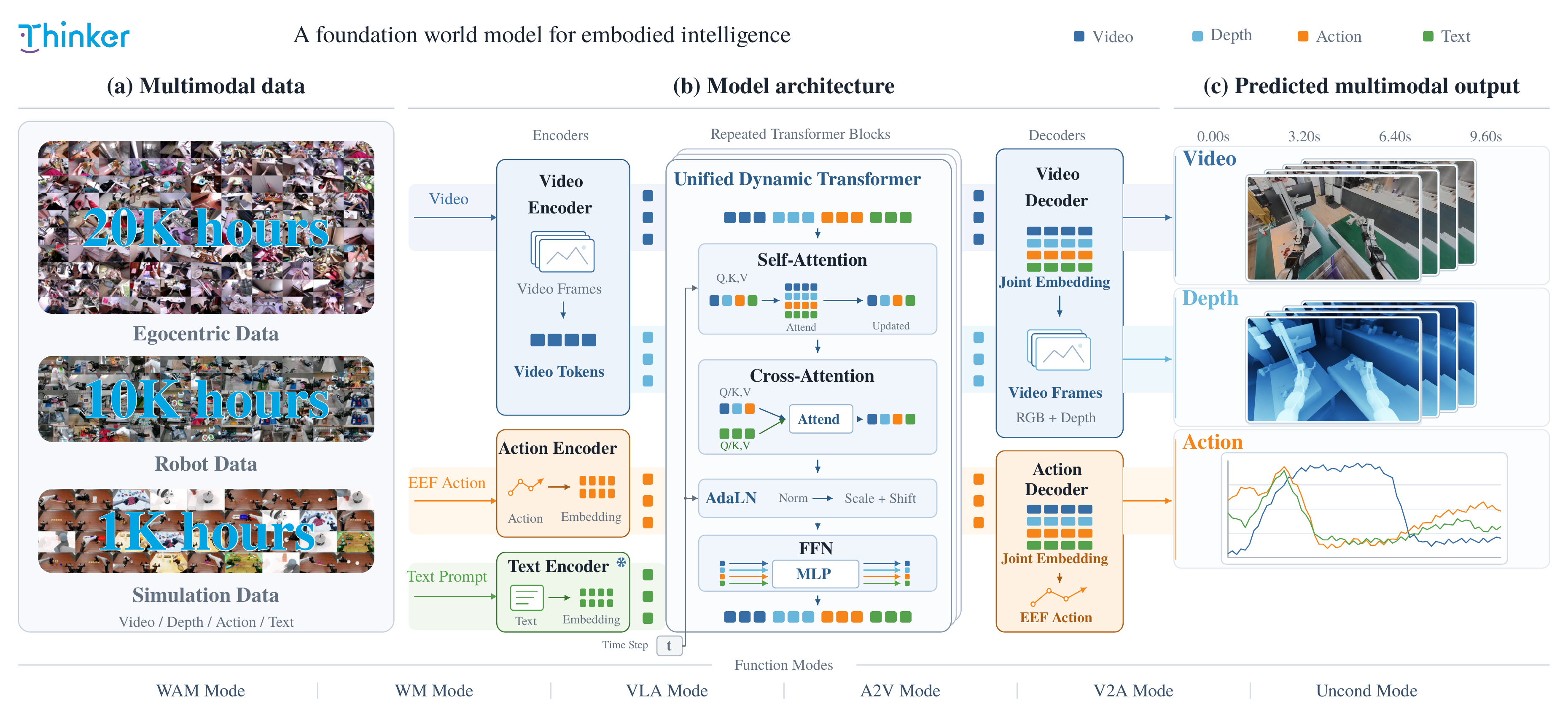} 
    \caption{\textbf{Thinker-WM overview.} Left: the pre-training mixture, which
combines egocentric, robot, and simulation data. Middle: every modality is lifted
into and decoded back from one unified representation by its own encoder and
decoder, while a shared Unified Dynamic Transformer (UDT) models the
future world within that representation using a single set of weights. Right: open-loop results on
real-world tasks, where video, depth, and action are generated jointly by the
same network.}
    \label{fig:overview}
\end{figure}

World models aim to learn how an environment evolves over time, enabling an agent to anticipate the consequences of its interventions rather than react only to the current observation. Early work developed compact latent simulators for planning and reinforcement learning \cite{worldmodels2018,dreamerv3}, while recent approaches increasingly leverage large-scale generative models with broader visual and temporal priors \cite{gaia1,genie,unisim,ivideogpt,cosmospredict}. For embodied intelligence, however, predicting future appearance alone is insufficient: the evolution of the world is simultaneously reflected in visual observations, physical geometry, executable actions, and task intent. Actions change the environment, the resulting state constrains subsequent actions, and task intent determines the desired evolution. This motivates world-action models (WAMs), which jointly model future world states and the actions that produce them \cite{motus,motubrain,uwm,dreamzero}.

Recent WAMs have made substantial progress by bringing video and action modeling into a common generative framework. Different approaches, however, place the coupling at different levels. Some systems compose modality-specific experts or maintain distinct processing pathways for different modalities \cite{motus,motubrain}. Others use a shared transformer and unify video and action through diffusion or noise configurations, while retaining separate modality channels \cite{uwm,diffusionforcing,reroll}. Omnimodal world models further broaden the set of modeled modalities, but may retain embodiment-specific projections around a shared backbone \cite{cosmos3}. Geometry has also been introduced into world-action modeling, for example through geometric-semantic prediction or lightweight spatial representations supervised by future depth \cite{geosemwam,wam4d,SplatPose}. These approaches demonstrate that combining complementary signals can improve embodied world modeling. Nevertheless, the modalities are often connected through interfaces, conditioning mechanisms, or modality-specific parameterizations rather than being expressed within a common representation that jointly carries future-world dynamics. This raises a fundamental question: where should multimodal unification take place in a general world-action model? We argue that the key is not merely to make multiple modalities available to the same network, but to let them evolve within a shared representation in which their common structure can be learned directly.

We therefore propose \textbf{Thinker-WM}, a unified multimodal world model that jointly models and generates RGB, depth, and action under task-level conditioning. The central design principle is representation-level unification. Each modality is first transformed by a modality-specific encoder into a shared intermediate representation. A common Transformer stage then models their joint evolution, while modality-specific decoders render the resulting representation back into RGB, depth, or executable action. The main unification therefore occurs inside the Unified Dynamic Transformer rather than only at its input or output interfaces. Within it, visual, geometric, and action tokens interact through common attention mechanism, while task-level representations provide structured guidance for future prediction. The resulting formulation treats appearance, geometry, action, and instruction as different manifestations of a coupled world process, while retaining modality-specific interfaces where the underlying physical representations necessarily differ. Figure~\ref{fig:overview} gives an overview of the data mixture, the architecture, and the open-loop use of the same network.

This representation-level formulation also allows geometry to become an explicit component of the generated world state. Thinker-WM represents depth using the same latent and spatiotemporal organization as visual generation, allowing geometric information to participate throughout the shared generative process. Geometric supervision therefore shapes the shared representation rather than being applied only through an auxiliary prediction branch. This makes physical structure directly available to the model alongside visual appearance and action, providing complementary information about the spatial relationships that govern embodied interaction. The objective is not to treat geometry as a complete solution to physical reasoning, but to make metric structure a native signal within the world model.

A unified representation is valuable only when the training distribution captures the diversity of embodied dynamics encountered in the real world. Thinker-WM is trained from a heterogeneous mixture of general video data, publicly available robot datasets, in-house robot data, and egocentric human video. General video provides broad visual and temporal priors; robot data provides action-grounded supervision across different embodiments, environments, and tasks; and egocentric video expands coverage of real-world interactions and human motion \cite{ego4d,egomimic,egodex,egoscale}. These sources differ substantially in sensor configuration, action representation, embodiment, and annotation format. Thinker-WM therefore maps heterogeneous robot actions into a common representation at the model boundary, allowing shared dynamics to be learned across embodiments while preserving robot-specific decoding. In this way, data diversity and architectural unification reinforce each other: diverse sources broaden the distribution of world dynamics, while the shared representation provides a common space in which those dynamics can be learned.

We evaluate Thinker-WM across embodied world-model benchmarks and a range of generative settings, including action-conditioned future generation, depth generation, long-horizon rollout, trajectory transfer, video-to-action generation, and instruction-conditioned generation. These experiments examine whether a shared multimodal representation can support world prediction, geometric modeling, and action generation within a single model. Thinker-WM ranked first on  EWMBench~\cite{ewmbench}, WorldOlympiad~\cite{2026WorldOlympiad} and WorldArena~\cite{2026WorldArena}, while additional experiments demonstrate its ability to support multiple world-model and world-action conditionals within the same parameterization.

The main contributions of this work are:
\begin{itemize}
\item \textbf{Representation-level multimodal unification.}
We introduce the Unified Dynamic Transformer in which RGB, depth, action, and task-level conditioning are lifted into a shared intermediate representation and transformed by common parameters, placing multimodal interaction inside the representation rather than only at the input or output interfaces.

\item \textbf{Geometry as part of the generative world state.}
We incorporate depth as a native generative modality that shares the latent and spatiotemporal structure of visual modeling and whose supervision directly shapes the shared representation, making physical geometry an explicit component of the modeled world dynamics.

\item \textbf{Scalable heterogeneous embodied learning.}
We combine general video, publicly available robot data, in-house robot data, and egocentric human video, together with a common action representation across embodiments, to provide complementary visual, geometric, and action supervision for scalable world-action modeling.

\end{itemize}

\vspace{2cm}
\section{Related Work}

\label{sec:related}

We view world models as generative models that learn how an environment
evolves under observations, conditions, and interventions. Their role has
gradually expanded from compact predictive simulators to large-scale
generative models with broad visual priors, and more recently to models
that jointly generate future observations and executable actions. From this
perspective, we organize prior work into three stages: simulator-oriented
world models, latent and generative world models, and world models that
generate actions alongside the future. We further discuss recent efforts to
introduce explicit geometric structure into this unified formulation, which
provides the closest context for Thinker-WM.

\subsection{Simulator-oriented World Models}

Early world models were developed primarily as learned simulators for
model-based planning and reinforcement learning. World Models
\cite{worldmodels2018} compress observations into a compact latent space,
model temporal dynamics with a recurrent mixture-density network, and
enable agents to train policies inside the model's imagined rollouts.
PlaNet and the subsequent Dreamer family~\cite{dreamerv3} further
demonstrate that a learned latent dynamics model can support planning and
behavior learning without requiring the policy to interact with the real
environment at every step. DreamerV3~\cite{dreamerv3}  in particular shows that a common
world-modeling recipe can operate across a broad collection of tasks and
domains under a largely fixed training configuration.

The central abstraction in this family is a predictive transition model,
\begin{equation}
\label{eq:wm-transition}
z_{t+1} \sim p_\theta(z_{t+1}\mid z_t,a_t),
\end{equation}
where $z_t$ is a learned latent state and $a_t$ specifies an intervention.
The model therefore functions primarily as a simulator: given a current
state and an action, it predicts a plausible future state on which planning
or control can be performed. This formulation is computationally efficient
and well suited to model-based decision making, but the learned latent
space is usually optimized around the environments and tasks encountered
during training. Consequently, the visual prior and open-world knowledge
available in large-scale image and video corpora are not directly inherited
by the dynamics model.

\subsection{Latent and Generative World Models}

Recent work increasingly treats the world model itself as a large-scale
generative model rather than a compact task-specific simulator. This shift
replaces manually designed or task-specific latent dynamics with generative
models capable of capturing richer visual structure and broader physical
priors. GAIA-1~\cite{gaia1}, for example, learns an action-conditioned
generative model of driving scenes from large-scale fleet data, while
Genie~\cite{genie} learns a controllable latent interface directly from
large collections of video without requiring explicit action annotations.
UniSim~\cite{unisim} further combines image, robotics, and navigation data
to construct a general-purpose simulator whose outputs can represent the
visual consequences of both high-level instructions and low-level controls.
iVideoGPT~\cite{ivideogpt} integrates action-conditioned video prediction
with visual planning and model-based reinforcement learning, while
Pandora~\cite{pandora} allows language-conditioned interventions throughout
a generated trajectory rather than restricting the condition to the
initial frame.

Large-scale video generation models provide another important foundation
for this transition. Rather than learning visual dynamics from scratch,
recent systems exploit pretrained video-generation backbones to inherit
strong priors over appearance, motion, and scene structure. The Cosmos
World Foundation Model~\cite{cosmospredict}, for example, positions a
pretrained generative world model as a reusable foundation for downstream
world-model construction and synthetic data generation. Xiaomi-Robotics-U0~\cite{xiaomiu0} similarly
explores broad multimodal generation and embodied scene synthesis,
positioning a world foundation model as both a generative world model and
a scalable source of downstream training data. These systems demonstrate
that large-scale visual pretraining can substantially broaden the
knowledge available to a world model beyond the distribution of a single
embodied environment.

Despite their generality, however, the action interface of these models is
often indirect. Actions may be represented as latent controls, natural
language commands, or environment-specific interfaces rather than as
executable physical robot actions. In other cases, the generated visual
trajectory is treated as a simulator for a separate downstream policy, or
the model serves as a data engine whose generated embodied data improves a
separately trained policy~\cite{xiaomiu0}.
Thus, the generative world model provides the predictive component of
control while the executable action model remains outside the generative
process. This separation motivates the next stage of world-model
development: jointly modeling the future world and the action that
produces it.

\subsection{World Models with Action}

This stage of the progression extends generative world modeling by
incorporating an executable action space into the same generative process as
future-world prediction. Instead of the sequential formulation
\begin{equation}
\hat{x}_{(t+1:t+H)}=g_\psi(x_t,c),
\qquad
a_t=h_\phi(\hat{x}_{t+1:t+H},x_t),
\end{equation}
where a video generator is followed by a separate inverse-dynamics model,
these approaches seek to model video and action jointly. The resulting
formulation allows the model to learn not only what the world may look like
after an intervention, but also which physical action is associated with
that evolution.

Motus~\cite{motus} is an early example of this unified formulation. It
combines pretrained experts for vision-language understanding, video
generation, inverse dynamics, and policy learning within a
Mixture-of-Transformers architecture, allowing several world-modeling and
control distributions to be represented in one system. To exploit
action-free video, Motus additionally introduces latent actions derived
from visual motion and uses a multi-stage curriculum to progressively
bridge video generation and robot control. This demonstrates that
large-scale video priors and action learning can be combined within a
single generative framework, while physical embodiment grounding is still
largely introduced at the later robot-specific stage.

MotuBrain~\cite{motubrain} further develops the unified formulation with a
three-stream architecture and a data pyramid spanning web video,
egocentric human video, and heterogeneous robot trajectories. In
particular, it treats egocentric video as video-only supervision rather
than requiring every video to have an explicit robot action label. Its
action representation is also expressed relative to the conditioning
frame, with relative position and rotation providing a more transferable
interface across robot configurations. These choices are closely related
to the setting considered in this work, where action-free egocentric video
and heterogeneous robot trajectories must coexist during training.
However, MotuBrain retains embodiment-specific physical scales rather
than explicitly registering the action statistics of different robots at a
shared model boundary, and its temporal instruction structure is primarily
episode-level rather than aligned to individual instruction segments.

A complementary line unifies video and action through diffusion noise
rather than through separate modality-specific architectures.
Diffusion Forcing~\cite{diffusionforcing} assigns independent noise levels
to different temporal tokens, enabling a single denoising model to express
multiple conditional distributions under different temporal masks.
UWM~\cite{uwm} applies this principle to robot learning by sharing a
transformer between image and action diffusion processes, allowing the
same model to operate as a policy, a world model, or a joint
video-action model depending on the masking configuration. Diffusion
ReRoll~\cite{reroll} further introduces asymmetric SNR-based reweighting
for image and action losses. These methods provide a flexible probabilistic
view of world modeling with action, although action and image remain separate
diffusion channels and their embodiment-specific representations are not
the primary focus.

Another direction expands the scope of modality unification. Cosmos~3
\cite{cosmos3} jointly models language, image, video, audio, and action
using a shared multimodal backbone, with domain-aware input and output
projections that carry each embodiment into a shared latent action space.
Ego and effector poses are encoded as frame-to-frame relative transforms,
while grasp states directly encode the current manipulation state, and
large-scale egocentric human video supplies additional pseudo-action
supervision. 

Recent work has additionally introduced explicit geometric structure into
this joint generative process. GeoSem-WAM~\cite{geosemwam} augments
joint world and action learning with geometric and semantic future prediction,
showing that structured spatial information can provide action-relevant
supervision beyond RGB appearance. WAM4D~\cite{wam4d} introduces
lightweight spatial register tokens supervised by future depth to inject
4D geometric priors into a video-action model, while removing the geometry
branch during action inference for efficiency. These results suggest that
metric geometry can provide a useful intermediate representation for
connecting visual dynamics and physical interaction. In these approaches,
however, geometry is primarily introduced as an auxiliary supervisory
signal or intermediate representation rather than as a modality that is
jointly generated throughout the same generative process.

DreamZero~\cite{dreamzero} provides the closest instance of the
pretrained-backbone route. It builds a robot foundation model on top of a
large image-to-video diffusion backbone and jointly denoises visual
trajectories and robot actions under a shared objective, showing that
preserving a strong visual generation prior improves downstream robot
control relative to vision-language-action baselines. Its remaining
boundaries are of a different kind: the instruction is supplied by a
frozen text encoder rather than produced by the generative process, the
model is pretrained separately for each embodiment, and its action target
stays at relative joint positions rather than a shared end-effector
representation. Zero-WAM~\cite{zero_wam} extends the same direction to
in-context human-video supervision, but keeps separate query, key, and
value projections, feed-forward networks, and output heads per modality,
so that video and action meet only inside the shared attention layers. The
two therefore share at different depths: one transformer with separated
decoders in DreamZero~\cite{dreamzero}, shared attention over separated parameters in
Zero-WAM~\cite{zero_wam}.

\subsection{Position of Thinker-WM}

The progression above moves from world models as compact simulators, to
large-scale generative models with transferable visual priors, and finally
to world models that jointly predict the future world and executable
actions. Thinker-WM follows the latter formulation while
extending it in three directions. First, it preserves a pretrained
flow-matching video-generation backbone and incorporates robot and
egocentric data without reducing action-free video to an action-labeled
proxy. Second, it treats appearance, metric geometry, and action as
co-evolving outputs of a shared intermediate representation rather than
using geometry only as an auxiliary training signal. Third, it registers
heterogeneous robot actions at the model boundary using a common relative
end-effector representation and per-robot statistics, allowing different
embodiments to share the same generative space while retaining exact
robot-specific decoding.

Architecturally, Thinker-WM therefore places the main unification inside the
UDT. Modality-specific encoders and decoders translate
video, geometry, action, and language into a common latent interface,
where shared attention and feed-forward transformations model their joint
evolution. The resulting formulation treats the world model not merely as
a simulator for a downstream policy, but as a unified generative model in
which visual dynamics, physical structure, and executable action are
different manifestations of the same underlying world process.

\vspace{2cm}
\section{Model Architecture}
\label{sec:model-architecture}
\subsection{Overview: Representation-Level Unified World Model}
\label{sec:architecture-overview}

A world model for embodied intelligence must capture more than the visual
appearance of the future. The evolution of an embodied environment is jointly
reflected in what the agent observes, the geometric structure of the scene, and
the actions that cause the observed changes. Thinker-WM therefore models RGB
video, metric depth, and robot action within a single generative framework,
rather than treating action generation as a downstream module of visual
prediction.

Given an observation $o_t$ and a task-level condition $c$, Thinker-WM models
the joint distribution of future visual, geometric, and action trajectories as
\begin{equation}
\label{eq:joint}
p\left(
v_{t+1:t+H},
d_{t+1:t+H},
a_{t+1:t+H}
\middle|
o_t,c
\right),
\end{equation}
where $v$, $d$, and $a$ denote RGB video, metric depth, and robot action,
respectively. The key design principle is \textbf{representation-level unification}: each generative modality is first encoded by a modality-specific encoder into an intermediate dynamic representation. These representations then feed into a Unified Dynamics Transformer (UDT) to participate in the shared dynamic evolution. The outputs are subsequently separated and decoded through
modality-specific generation head.

This design satisfied two requirements that are often conflated in multimodal
models. On the one hand, RGB, depth, and action have fundamentally different
input statistics and output semantics and therefore require dedicated
interfaces. On the other hand, their future evolution is physically coupled:
actions alter the environment, visual observations reveal the consequences,
and metric geometry constrains both perception and interaction. 

Thinker-WM is initialized from a pretrained Wan2.2 TI2V-5B video
generation backbone and extends it with dedicated depth and action pathways.
A frozen Qwen3-VL-2B~\cite{bai2025qwen3vltechnicalreport} model provides task-level understanding, which is
projected into the model as a structured conditioning representation rather
than treated as a fourth generative stream. The resulting information flow can
be summarized as
\begin{equation}
\label{eq:architecture-flow}
(o_t,c)
\rightarrow
\left\{
\mathcal{E}_v,\mathcal{E}_d,\mathcal{E}_a,
\mathcal{E}_{\mathrm{l}}
\right\}
\rightarrow
\mathcal{U}
\rightarrow
\left\{
\mathcal{D}_v,\mathcal{D}_d,\mathcal{D}_a
\right\}
\rightarrow
(\hat v,\hat d,\hat a),
\end{equation}
where $\mathcal{E}_v$, $\mathcal{E}_d$, and $\mathcal{E}_{a}$ are the modality-specific
generative encoders, $\mathcal{E}_{\mathrm{l}}$ produces the task-level understanding representation, $\mathcal{U}$ denotes the unified dynamics transformer, and $\mathcal{D}_v$, $\mathcal{D}_d$, and $\mathcal{D}_a$ decode
the corresponding future modalities. The following sections describe the latent interfaces, the
construction of the unified dynamics transformer, and the generative
parameterization in detail.

\begin{table}[htbp]
\centering
\caption{Structure and weight origin of the three-stage architecture and
auxiliary branches. Pretrained visual blocks are inherited from Wan2.2
TI2V-5B and remain trainable, while the VAE and VLM are frozen.}
\label{tab:three-stage}
\small
\begin{tabularx}{\linewidth}{@{}lclX@{}}
\toprule
\textbf{Component} & \textbf{Blocks} & \textbf{Weight origin} & \textbf{Role} \\
\midrule

Visual encoder $\mathcal{E}_{v,d}$
& 8 blocks
& Inherited
& Encode RGB and physical-fidelity tokens. \\

RGB UDT $\mathcal{U}$
& 14 blocks
& Inherited
& Model appearance and interaction. \\

Physical-fidelity UDT $\mathcal{U}$
& 14 blocks
& Inherited
& Model geometry and interaction. \\

Visual decoder $\mathcal{D}_{v,d}$
& 8 blocks
& Inherited
& Decode RGB and physical-fidelity outputs. \\

Action encoder $\mathcal{E}_a$
& 8 blocks
& Initialized
& Encode action tokens. \\

Action decoder $\mathcal{D}_a$
& 8 blocks
& Initialized
& Decode executable actions. \\

Understanding \texttt{und}
& 4 MLP blocks
& Initialized + Qwen3-VL
& Provide task and planning guidance. \\

\bottomrule
\end{tabularx}
\end{table}

\subsection{Multimodal Representation}
\label{sec:backbone-split}

\subsubsection{Language Representation}
\label{sec:conditioning}

Language is represented as a task-level conditioning sequence rather than a
fourth generative stream. 
The structured prompt is processed by a frozen Qwen3-VL-2B~\cite{bai2025qwen3vltechnicalreport}. Its final hidden states are truncated
to at most 1000 tokens and projected into the model width through a linear
bridge followed by four residual MLP blocks, producing the conditioning
sequence $h_{und}$. 

Instructions from heterogeneous datasets are
normalized into a structured three-field prompt containing \textit{Task},
\textit{Scene}, and \textit{Action}. The fixed schema separates the task
objective from the scene description and the intended operation, reducing
variation in prompt format across data sources and providing a consistent
semantic structure for the language encoder.
Prompt structure is further regularized during training. Redundant
occurrences of the field are merged, the order of the three fields is
randomly shuffled, and random field dropout is applied. Evaluation uses a fixed \textit{Task--Scene--Action}
order. These transformations encourage the model to rely on semantic
content rather than field position or the presence of a complete schema.


\subsubsection{Visual Representation}
\label{sec:visual-representation}

RGB frames and depth maps are both compressed by the frozen Wan2.2
VAE into the same latent topology. At $640\times480$ resolution, a window of
49 frames is mapped to a latent grid of $13\times30\times40$, and
patchification with kernel $(1,2,2)$ produces
$13\times15\times20=3900$ tokens for each visual stream. Using the same VAE
and spatiotemporal organization allows RGB and depth to occupy compatible
latent coordinates while retaining distinct modality semantics.

\begin{figure}[t]
    \centering

    \begin{minipage}[t][6.7cm][t]{0.49\textwidth}
        \centering
        \vspace{0pt}

        \includegraphics[
            width=\linewidth
        ]{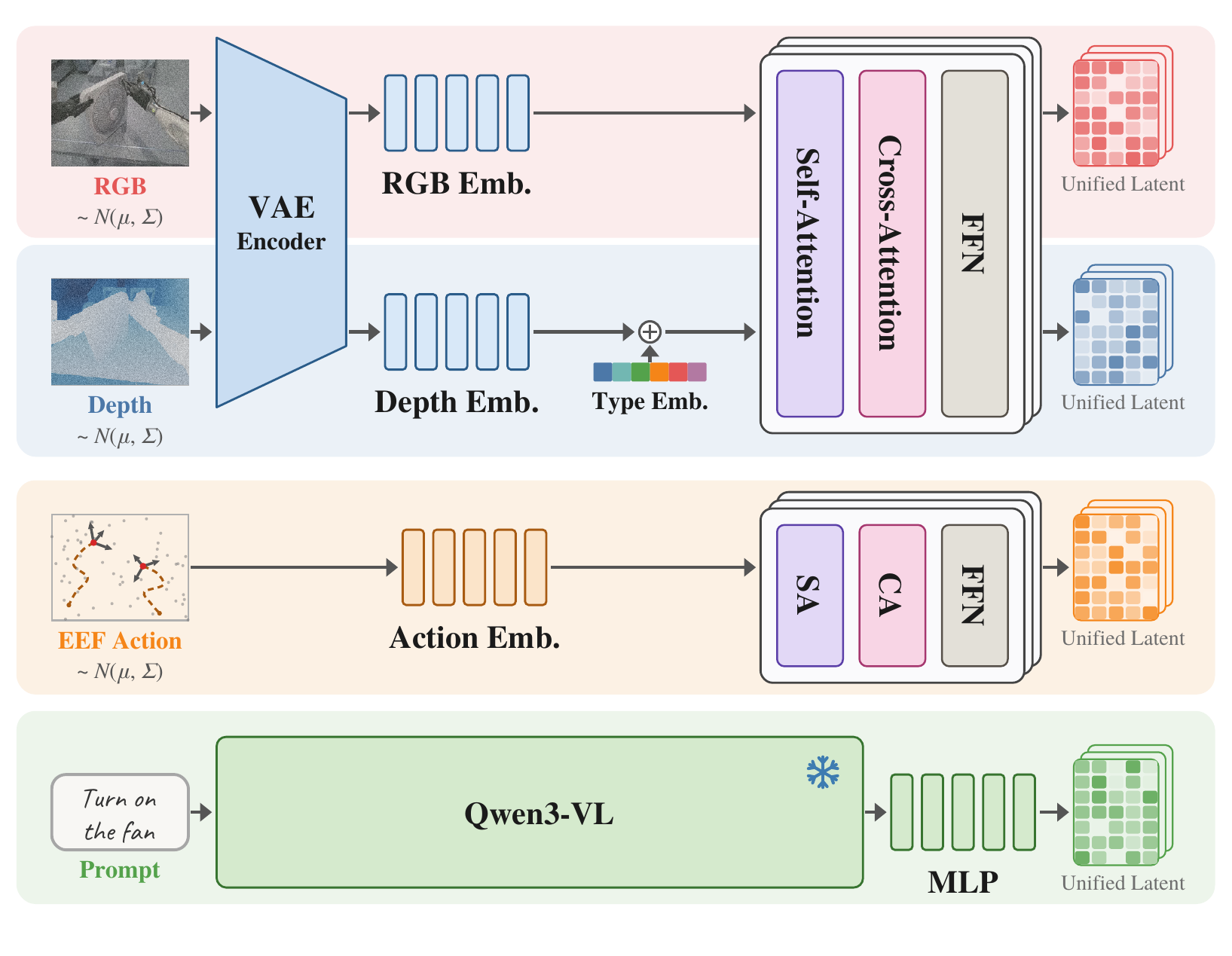}

        \vfill

        \textbf{(a)} Encoder stage
    \end{minipage}
    \hfill
    \begin{minipage}[t][6.7cm][t]{0.49\textwidth}
        \centering
        \vspace{0pt}

        \includegraphics[
            width=\linewidth
        ]{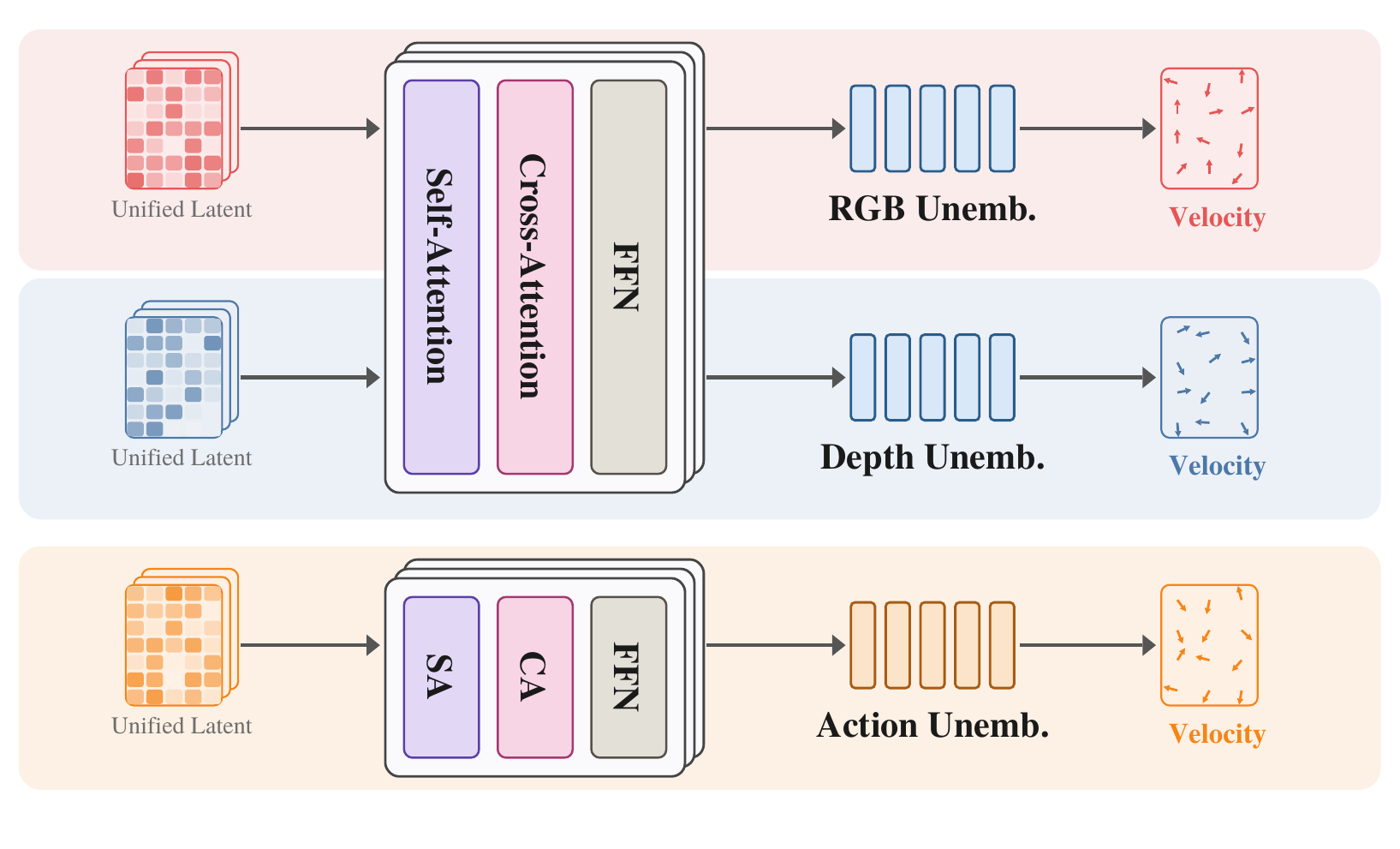}

        \vfill

        \textbf{(b)} Decoder stage
    \end{minipage}

    \vspace{6pt}

\caption{\textbf{Multimodal encoding and decoding pipelines.}
RGB video, metric depth, and action are mapped into the unified latent
space through modality-specific encoders and processed back into their
respective output spaces through modality-specific decoders. The frozen
Qwen3-VL-2B converts the structured prompt into the conditioning sequence
$h_{\texttt{und}}$, while robot identity is provided through learnable
tokens $r$. The decoder stage predicts the flow-matching
velocity field for each modality rather than directly producing decoded
samples.}
    \label{fig:multimodal-pipelines}
\end{figure}

The two visual modalities nevertheless use separate input projections.
RGB tokens are mapped directly into the model width, while depth tokens use
their own patch embedding followed by a learnable stream embedding that
distinguishes geometric information from appearance. After multimodal
dynamics modeling, the two streams are mapped back through modality-specific
output projections. Thus, RGB and depth share a common latent topology, but
their input and output representations remain explicitly separated where
their underlying statistics differ. Figure~\ref{fig:multimodal-pipelines}
illustrates the resulting visual encoding paths.

\subsubsection{Unified Action Representation}
\label{sec:unified-action}
Robot actions are substantially more heterogeneous than visual inputs:
different embodiments expose different action dimensions, coordinate frames,
control conventions, and actuator interfaces. A shared generative model
therefore requires an action representation that is independent of the raw
dimension of any particular robot. Thinker-WM maps every embodiment into a
fixed unified action space of $U=20$ dimensions, organized as dual-arm
delta end-effector positions, rotations, and gripper states.
For robot $r$ with raw action $a\in\mathbb{R}^{d_r}$, we define a
semantic slot map
\[
\pi_r:\{1,\ldots,U\}
\rightarrow
\{1,\ldots,d_r\}\cup\{-1\},
\]
where $\pi_r(u)$ specifies which raw action dimension corresponds to
unified slot $u$, and $\pi_r(u)=-1$ indicates that the robot does not
have that degree of freedom. The unified action and its validity mask are
then defined as
\begin{equation}
\label{eq:scatter}
\tilde{a}_u =
\begin{cases}
a_{\pi_r(u)}, & \pi_r(u)\ge 0,\\
0, & \pi_r(u)=-1,
\end{cases}
\qquad
m_u =
\begin{cases}
1, & \pi_r(u)\ge 0,\\
0, & \pi_r(u)=-1.
\end{cases}
\end{equation}
Here, $m_u$ indicates whether unified dimension $u$ is valid for robot
$r$. Invalid dimensions are neither noised nor included in the training
loss, so the zero-filled entries do not become artificial action targets.

We represent all robot actions using a unified relative end-effector (EEF)
representation. For each end effector, the position and orientation at time
$t$ are expressed relative to its state at the conditioning frame:
\begin{equation}
\label{eq:relative}
\Delta p_t=
p_t-p_0,
\qquad
R_t^{\mathrm{rel}}=
R_0^\top R_t .
\end{equation}
Here, $\Delta p_t$ denotes the relative EEF displacement and
$R_t^{\mathrm{rel}}$ the relative EEF rotation, represented using the
continuous 6D rotation parameterization. Anchoring each trajectory to its own
conditioning frame removes dependence on the global coordinate frame and
makes action representations more comparable across scenes, camera poses,
and embodiments. 

Action normalization is applied at the model boundary using statistics
computed separately for each robot. Gripper and dexterous-hand dimensions
are scaled to $[-1,1]$ while the
remaining dimensions retain their physical units. The visual and action streams are temporally aligned with a fixed \(6{:}1\) correspondence, with six action steps per video frame. A temporal convolution with kernel and stride
6 compresses the action sequence into 48 tokens of width 3072, while the
action decoder expands each token back to $6\times20$ action values. The
resulting action tokens therefore share the temporal horizon of the visual
latent sequence while remaining compact enough for joint processing.

Embodiment identity is provided through a small set of learnable robot
tokens. For each robot, eight learnable tokens are prepended to the
language conditioning sequence to form the cross-attention context of the
encoder and decoder:
\begin{equation}
\label{eq:robot-tokens}
c = [\,r_{1:8}\;;\;h_{und}\,].
\end{equation}
Placing robot identity in the conditioning context rather than in the main
generative sequence preserves the sequence structure shared across
embodiments.

Together, these three representations establish the model boundary:
visual streams preserve a shared spatiotemporal latent topology, heterogeneous
robot actions are registered into a common 20-dimensional action space, and
language is converted into a structured conditioning sequence. These
representations are then passed to the Unified Dynamics Transformer, where
their interactions and joint temporal evolution are modeled explicitly.

\subsection{Unified Dynamics Transformer}
\label{sec:udt}

The Unified Dynamics Transformer (UDT) is the central multimodal dynamics
modeling stage of Thinker-WM. The encoded RGB, depth, and action tokens are
concatenated into a joint generative sequence, while the language
representation provides task-level conditioning through cross-attention.
The UDT therefore operates on a common latent sequence while preserving
modality-specific representations at its input and output boundaries.

Given the modality-specific latent representations
$z_v^0$, $z_d^0$, and $z_a^0$, the unified dynamics stage is written as
\begin{equation}
\label{eq:unified}
z'
=
\mathcal{U}
\left(
[\,z_v^0\,;\,z_d^0\,;\,z_a^0\,],
h_{\texttt{und}}
\right),
\end{equation}
where $[\,\cdot\,;\,\cdot\,]$ denotes concatenation along the token
dimension and $h_{\texttt{und}}$ is the task-level conditioning sequence.
The output $z'$ is subsequently separated into modality-specific streams
for RGB, depth, and action decoding.

\begin{figure}[!ht]
    \centering
    \includegraphics[width=0.8\textwidth]{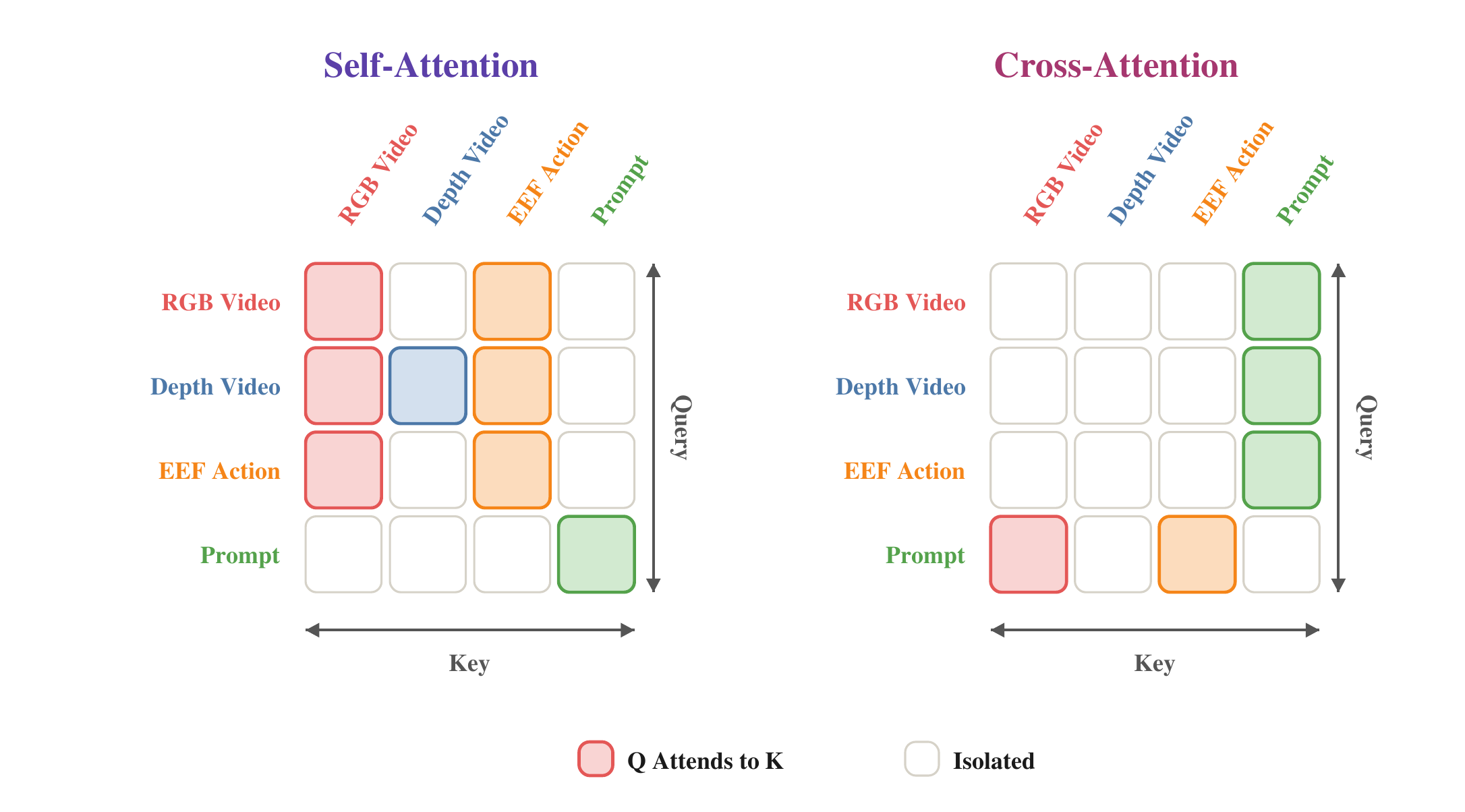}
    \caption{\textbf{Attention pattern of the UDT during pretraining.}
    Self-attention provides cross-modal interaction among RGB, depth, and
    action, while cross-attention provides the dedicated task-conditioning
    path from the prompt representation. Depth interacts asymmetrically with
    the other generative streams during pretraining.}
    \label{fig:attention-flow}
\end{figure}

The attention structure during pretraining is shown in
Figure~\ref{fig:attention-flow}. RGB and action form the main generative
interaction path, allowing the two streams to exchange information through
self-attention. Depth is asymmetric: it reads from the visual and action
streams but does not feed back into them. This one-way coupling allows
geometric supervision to shape the shared representation while preventing
imperfect depth supervision from directly perturbing the pretrained visual
and action pathways. The language tokens are isolated from self-attention
and provide task-level conditioning through the dedicated cross-attention
path shown in the figure.

\begin{figure}[!ht]
    \centering
    \includegraphics[width=0.8\textwidth]{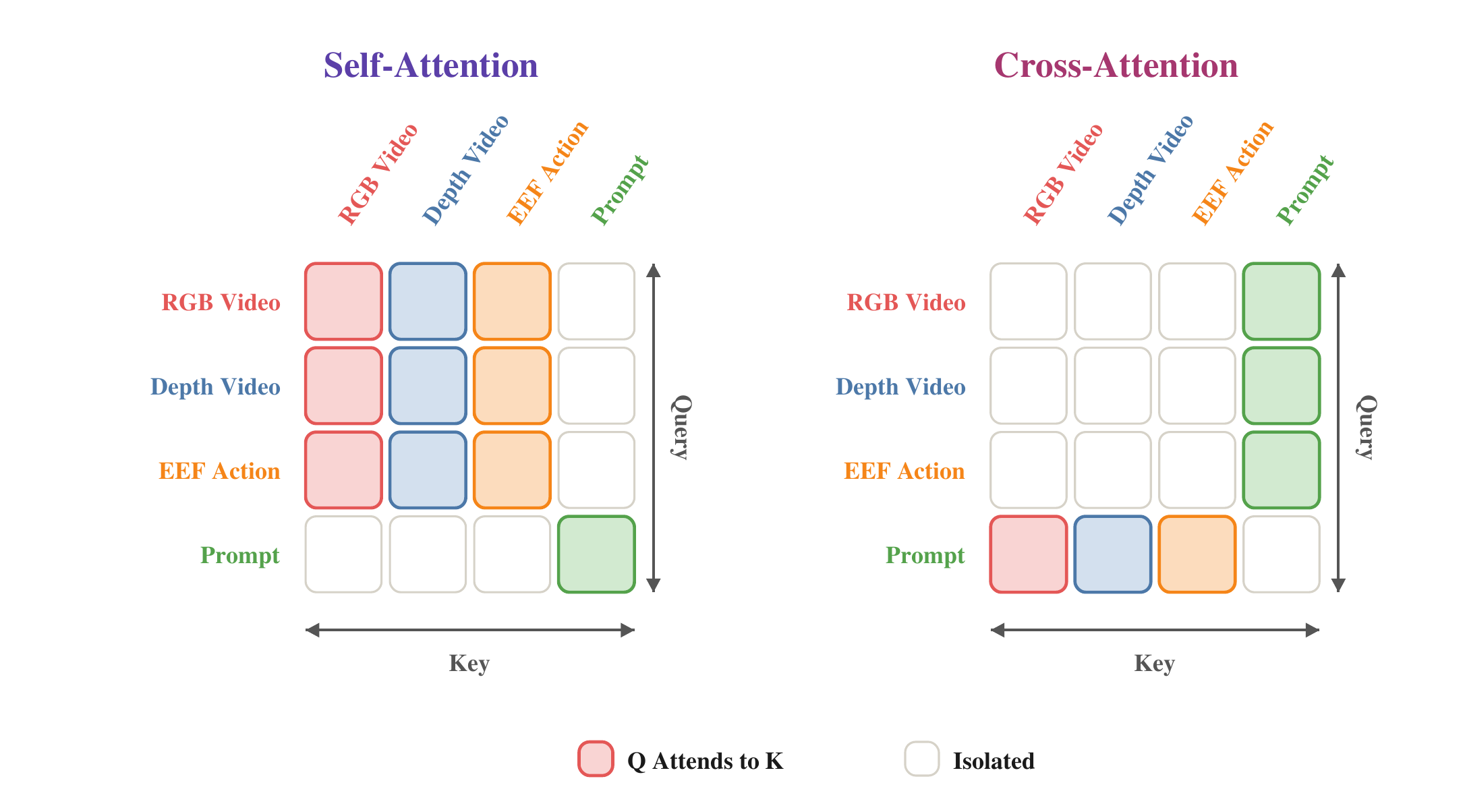}
    \caption{\textbf{Attention pattern of the UDT during post-training.}
    Compared with pretraining, the geometric stream is opened to
    bidirectional interaction with RGB and action, enabling multimodal
    refinement under reliable depth supervision.}
    \label{fig:attention-flow-sft}
\end{figure}

In the post-training stage, 
RGB and action tokens are also allowed to attend to depth tokens
(Figure~\ref{fig:attention-flow-sft}), enabling bidirectional interaction
among RGB, depth, and action. This changes the role of geometry from a one-way supervisory
signal into a bidirectional source of multimodal refinement, allowing
appearance, geometry, and action to jointly adapt during the final training
stage. The original language representation
$h_{\texttt{und}}^{0}$ is first provided to the encoder, and the UDT
maintains both the original and the layer-updated language representations.
For UDT block $k$, the cross-attention keys and values are constructed as

\begin{equation}
\label{eq:compact-kv}
K_k
=
W_K[\,h_{\texttt{und}}^{0};\,h_{\texttt{und}}^{k}\,],
\qquad
V_k
=
W_V[\,h_{\texttt{und}}^{0};\,h_{\texttt{und}}^{k}\,].
\end{equation}
This preserves the original task condition while allowing the language
representation to adapt to the evolving multimodal state. 

The UDT further uses modality-aware positional and timestep conditioning.
RGB tokens use 3D rotary position encoding, language tokens use
one-dimensional positional encoding, and action tokens use temporal-only
rotary encoding with a frequency scaling matched to the fixed temporal
alignment between video and action. A shared timestep embedding provides
the adaLN modulation for the generative streams using their respective
flow times.
\subsection{Function-Oriented Training}
\label{sec:function-training}

The shared multi-modal representation makes it possible to use the same
parameterization under different combinations of observed and generated
modalities. In particular, RGB, depth, and action are modeled within the
same UDT, while a modality can either be generated from noise, provided as a
clean condition, or omitted from the computation. This flexibility allows
the same model to realize joint world--action generation, visual world
prediction, policy inference, and cross-modal generation without introducing
separate networks.

\begin{table}[h]
  \centering
  \caption{\textbf{Function-oriented training modes.}
  Each training mode corresponds to one function of the
  shared model.}
  \label{tab:function-modes}
  \begin{tabular}{@{}lll@{}}
    \toprule
    \textbf{Mode} & \textbf{Training Detail} & \textbf{Role}\\
    \midrule
    WAM & joint denoising of visual streams and action
        & joint generation \\
    WM & action branch omitted
       & visual world model \\
    VLA & first video frame kept clean
        & policy on the current observation \\
    A2V & clean action injected, prompt dropped
        & action-conditioned visual generation \\
    V2A & clean video injected, prompt dropped
        & video-conditioned action generation \\
    Uncond & prompt dropped, joint denoising
           & unconditional branch for guidance \\
    \bottomrule
  \end{tabular}
\end{table}

\begin{figure}[!ht]
    \centering
    \includegraphics[width=\textwidth]{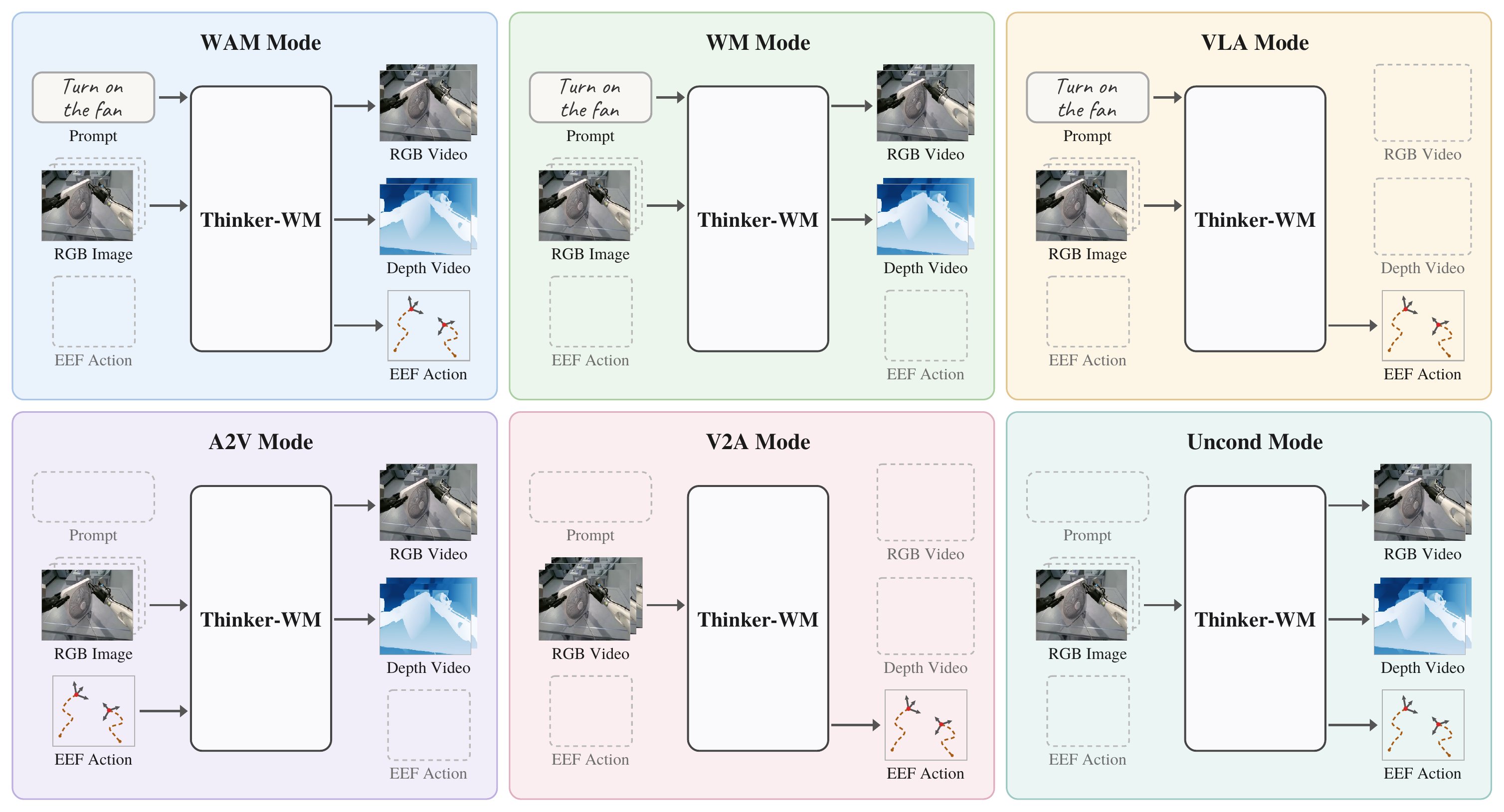}
    \caption{\textbf{Function modes of Thinker-WM.}
   Each panel
    shows the modality configuration of one function, with dashed outlines
    indicating omitted inputs or outputs.}
    \label{fig:function-modes}
\end{figure}

Rather than relying on random modality
dropout to implicitly cover these conditionals, Thinker-WM explicitly
assigns a training configuration to each function mode, as illustrated in  Table~\ref{tab:function-modes} and Figure~\ref{fig:function-modes}. The function modes include joint generation (WAM), visual world modeling (WM), action generation from the current observation (VLA), action-conditioned
visual generation (A2V), video-conditioned action generation (V2A), and the
unconditional branch required for classifier-free guidance.

Conditional generation is implemented by assigning zero flow time to the
conditioning modality. Under the flow-matching path in
Eq.~\eqref{eq:noisy_target_35}, setting $\sigma=0$ yields
\begin{equation}
\label{eq:gt-inject}
x_{\sigma=0}=x.
\end{equation}
The conditioned modality is therefore passed through its original encoder as
a clean observation, while its prediction loss is excluded from the
objective. The same construction is used for action-conditioned visual
generation, video-conditioned action generation, and VLA, where the first
video frame is provided as the clean observation.

For each training mode, omitted modalities are excluded from both the
forward computation and their associated loss terms. The resulting
optimization objective therefore matches the corresponding inference
configuration: conditioned modalities are held fixed as observations,
generated modalities are supervised against their target distributions, and
omitted modalities are absent from the computation graph. This explicit
alignment between training and inference allows a single
parameterization to support multiple world-model and cross-modal generation
settings.

\subsection{Flow-Matching Parameterization}
\label{sec:flow-matching}

Thinker-WM applies a unified velocity-field parameterization to every generative modality. For modality \(m\in\{v,d,a\}\), let \(x_m\) denote its clean representation, \(\epsilon_m\sim\mathcal{N}(0,I)\), and \(\sigma_m\in[0,1]\). The noisy state and target velocity are
\begin{equation}
x_{m,\sigma_m}=(1-\sigma_m)x_m+\sigma_m\epsilon_m,\qquad
u^\star_m=\epsilon_m-x_m.
\label{eq:noisy_target_35}
\end{equation}
The model predicts \(\hat{u}_m=v_{\theta,m}(x_{m,\sigma_m},\sigma_m,c)\) for each generative stream, where \(c\) denotes the corresponding multimodal conditioning context.

RGB and depth follow the same flow trajectory and therefore share a common noise schedule, while the action stream uses a separate schedule:
\begin{equation}
\label{eq:shift}
\sigma(s)
=
\frac{\mathrm{sh}\cdot s}
{1+(\mathrm{sh}-1)\cdot s},
\qquad
s\sim\mathrm{Uniform}\{1,\dots,1000\},
\end{equation}
with \(\mathrm{sh}=6.0\) for RGB and depth and \(\mathrm{sh}=1.0\) for action. The modality-specific schedules are retained at inference so that the sampling trajectories remain consistent with the corresponding training distributions.

At inference, the learned velocity fields are integrated using 50 steps for RGB and depth and 10 steps for action. Classifier-free guidance is applied only to the visual streams, while the action branch is sampled without guidance to avoid instability in the generated action trajectories.

\subsection{Training Objective}
\label{sec:training_objective}

Given the velocity-field parameterization above, the training loss for each generative modality is the standard flow-matching regression objective:
\begin{equation}
\mathcal{L}_m(\theta)=
\mathbb{E}_{x_m,\epsilon_m,\sigma_m}
\left[
\left\|
v_{\theta,m}(x_{m,\sigma_m},\sigma_m,c)-u^\star_m
\right\|_2^2
\right].
\label{eq:modality_loss_36}
\end{equation}

For the action stream, the unified action space has \(U=20\) dimensions with a validity mask \(m_u\) that retains only the degrees of freedom available to the robot. The action loss is
\begin{equation}
\mathcal{L}_a(\theta)=
\mathbb{E}
\left[
\frac{
\sum_{u=1}^{U}m_u
\left\|
\hat{u}_{a,u}-u^\star_{a,u}
\right\|_2^2
}{
\sum_{u=1}^{U}m_u+\varepsilon
}
\right],
\label{eq:action_loss_36}
\end{equation}
where \(\varepsilon\) is a small numerical stabilizer. Invalid action dimensions are neither noised nor included in the loss; when the loss is computed in the action-token space, the validity mask is expanded to the corresponding token dimensions. The modality-specific noise schedules in Section~\ref{sec:flow-matching} are used when evaluating the corresponding \(\mathcal{L}_m\).

At training time, the current function mode \(r\) partitions the modalities into a generated set \(\mathcal{G}(r)\), a conditioned set \(\mathcal{C}(r)\), and an omitted set \(\mathcal{O}(r)\). The total objective is
\begin{equation}
\mathcal{L}_{\mathrm{total}}(\theta)=
\sum_{m\in\mathcal{G}(r)}
\lambda_m\mathcal{L}_m(\theta;r),
\label{eq:total_loss_36}
\end{equation}
where \(\lambda_m\) balances the modality scales or effective token counts. For \(m\in\mathcal{C}(r)\), setting \(\sigma_m=0\) feeds the modality as a clean observation while excluding its prediction loss; for \(m\in\mathcal{O}(r)\), the modality is absent from both the forward computation and the loss. Consequently, the six function modes---WAM, WM, VLA, A2V, V2A, and Uncond---share the same loss form and differ only in which modalities are generated, conditioned, or omitted. The modes are sampled with fixed proportions determined by the global step index, ensuring consistency across data-parallel ranks. 
\section{Data}
\label{sec:Data}

\begin{figure}[htbp]
    \centering
    \includegraphics[width=\textwidth]{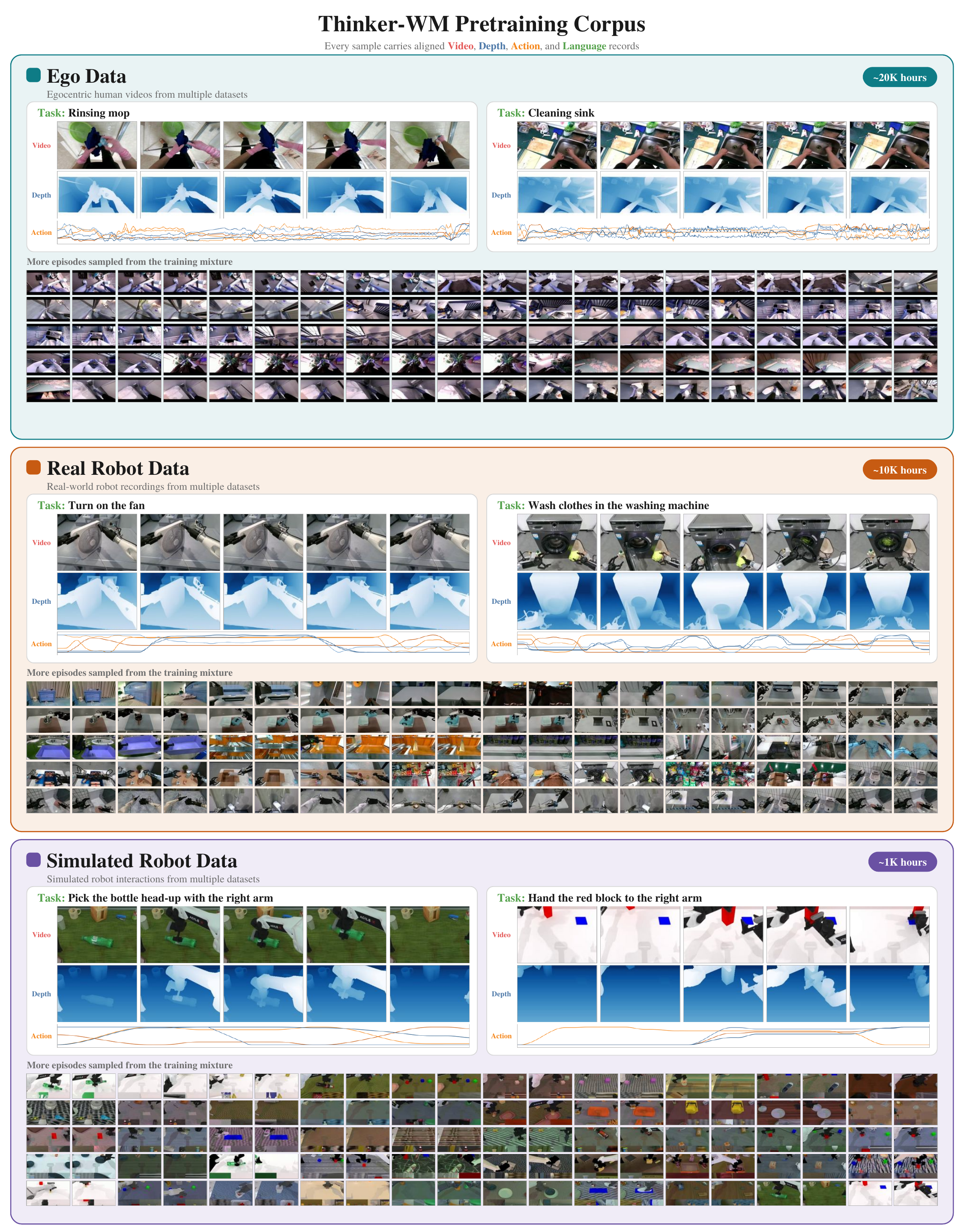}
    \caption{\textbf{Training data composition}. For each category, two example episodes are shown with five video frames, their aligned depth labels, and the corresponding action curves; task text serves as the language record. The grids show additional episodes sampled from the training mixture. Depth is obtained by stereo matching for egocentric video, predicted by a monocular teacher for real robot video, and rendered exactly in simulation.}
    \label{fig:data-composition}
\end{figure}

Our pretraining corpus aggregates approximately \textbf{20,000} hours of high-quality egocentric human video and \textbf{10,000} hours of robot data, including around \textbf{4,000} hours with aligned depth. We organize the corpus into in-house data, internet data, and depth data. Figure~\ref{fig:data-composition} summarizes its composition.

\subsection{In-House Data}
\label{sec:in-house-data}

In-house data provide coverage of our target embodiments, environments, and task distributions. It includes high-quality egocentric human video collected internally, spanning household, commercial, retail, and industrial scenarios. These egocentric recordings are captured with head-mounted stereo devices at 30Hz, and each frame includes hand keypoints and head pose. We further organize this egocentric corpus into 10 major categories and 20 subcategories, with the data evenly distributed across these categories. This egocentric data supplies scalable task, scene, object, and long-horizon activity diversity that robot teleoperation cannot economically reach. We also integrate high-quality real-robot data collected from UBTech robots, which provides paired visual observations and executable actions for learning practical robot dynamics and embodiment-specific action representations.

\subsection{Internet Data}
\label{sec:internet-data}

Internet data consist of open-source datasets collected from the web. General video data, including OpenVid~\cite{nan2024openvid}, cover diverse scenes, objects, activities, and visual dynamics, providing broad visual and temporal priors beyond the distribution of embodied datasets. Open-source robot datasets, including AgiBot~\cite{agibotworld} and RoboMIND~\cite{wu2025robomind}, cover diverse embodiments, environments, and manipulation tasks, offering paired visual observations and actions for learning embodied dynamics and cross-embodiment representations. RoboTwin~\cite{robotwin} provides simulated manipulation data. Egocentric human video sources, including EgoVerse~\cite{punamiya2026egoverseegocentrichumandataset} and Ego4D~\cite{ego4d}, supply scalable task, scene, object, and long-horizon activity diversity. The sources listed here are representative rather than exhaustive; additional public datasets are also included in the pretraining mixture.

\subsection{Depth Data}
\label{sec:depth-data}

\begin{figure}[h]
    \centering
    \includegraphics[width=\textwidth]{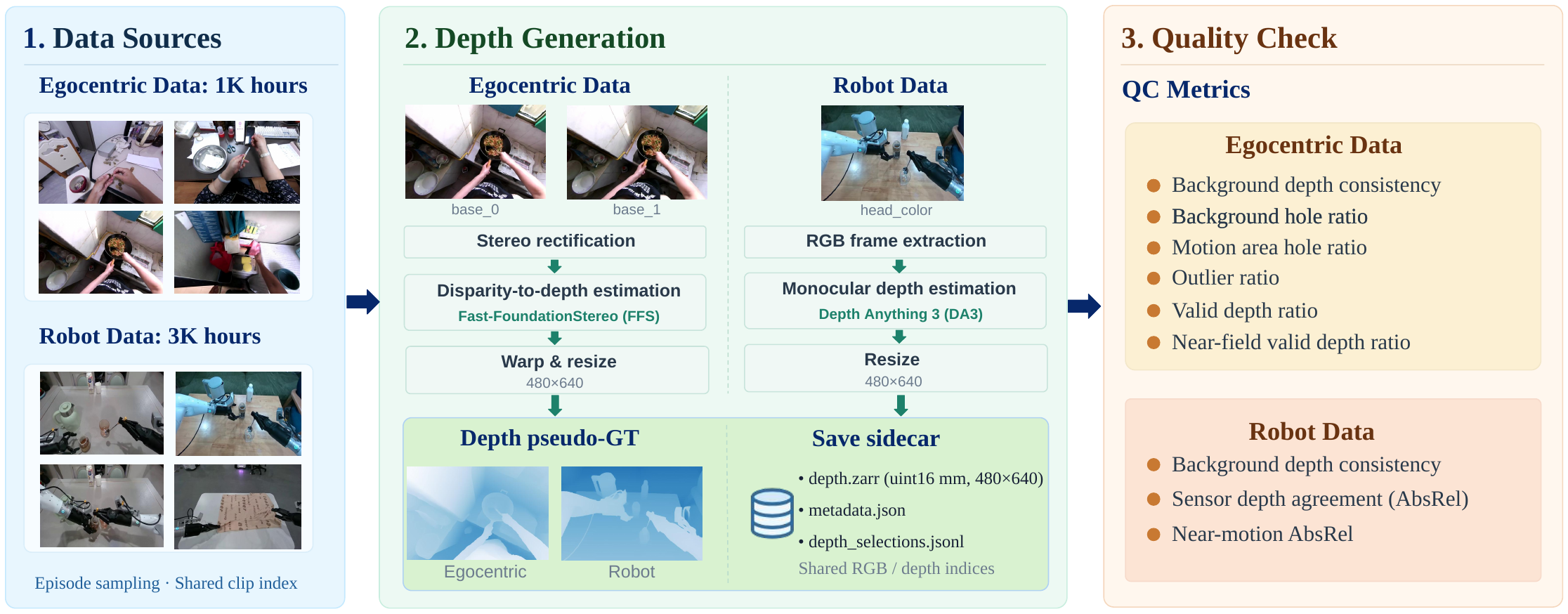}
    \caption{\textbf{Depth data pipeline}. We generate depth pseudo-labels for egocentric and robot videos, filter low-quality clips, and align the retained depth with RGB frames for world model training.}
    \label{fig:depth-pipeline}
\end{figure}

To complement RGB observations with explicit geometric information, we augment egocentric and robot videos with depth pseudo-labels. We use Fast-FoundationStereo~\cite{ffs} to estimate depth from egocentric stereo pairs and DepthAnything3~\cite{da3} to predict depth from monocular robot videos. The resulting depth maps are aligned with the original RGB views and resized to a common resolution. We apply clip-level quality control based on temporal consistency, depth validity, and agreement with available sensor depth, discarding clips with unreliable predictions. During training, we sample synchronized RGB and depth sequences, apply logarithmic depth encoding, and encode both modalities with the same frozen VAE. Depth is then learned as an additional generative stream using flow matching, providing geometric supervision for modeling spatial structure and physical interactions.

\newpage

\vspace{2cm}
\section{Infrastructure}
\label{sec:infrastructure}

\subsection{Overview}
\label{sec:infra-overview}

\begin{figure}[!ht]
    \centering
    \includegraphics[width=\textwidth]{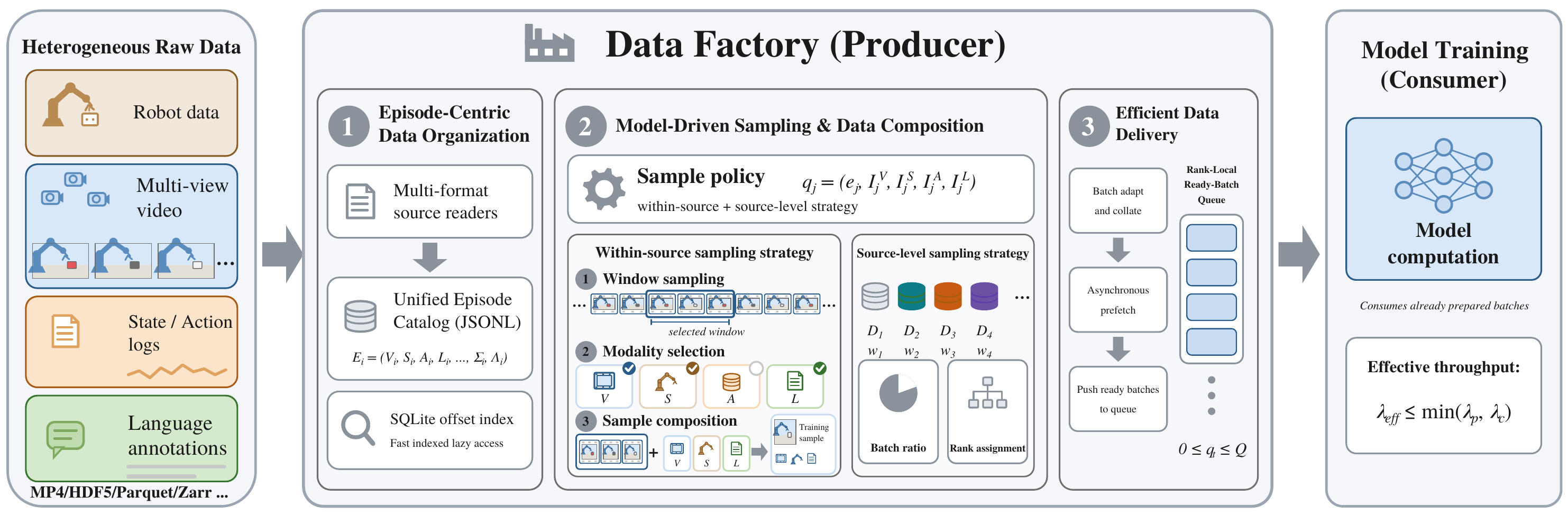}
\caption{
Overview of the Data Factory producer--consumer pipeline.
Heterogeneous multimodal data are organized into episode-centric catalogs
\(E_i=(V_i,S_i,A_i,L_i,\Sigma_i,\Lambda_i)\) and accessed lazily through
a SQLite offset index.
The model specifies each training sample as
\(q_j=(e_j,I_j^V,I_j^S,I_j^A,I_j^L)\), including within-source window and
modality selection, while source-level sampling controls the mixture of data
sources across batches and ranks.
The resulting samples are adapted, collated, and asynchronously prefetched into
a rank-local ready-batch queue with occupancy \(0\le q_t\le Q\), from which the
training process consumes prepared batches.
The effective pipeline throughput is bounded by
\(\lambda_{\mathrm{eff}}\le\min(\lambda_p,\lambda_c)\), where
\(\lambda_p\) and \(\lambda_c\) denote the producer and consumer rates.
}
    \label{fig:infra-overview}
\end{figure}

World-model training relies on heterogeneous data sources that were not originally collected for joint use. These sources span robot trajectories from diverse embodiments, simulation rollouts, public video, and egocentric human recordings, each with distinct storage formats, modalities, temporal structures, action spaces, and annotation conventions. The \textbf{Data Factory} separates a data layer that maintains factual records from a training layer that specifies sampling and composition policies. This decoupling allows either side to evolve without rebuilding the other. Figure~\ref{fig:infra-overview} illustrates the producer--consumer architecture.

The fundamental unit is the \textit{Episode}, which carries its own modalities, index spaces, and physical locations:
\begin{equation}
\label{eq:infra-episode}
E_i = \left(V_i, S_i, A_i, L_i, \Sigma_i, \Lambda_i\right),
\end{equation}
where \(V_i, S_i, A_i, L_i\) denote video, state, action, and language modalities, and \(\Sigma_i, \Lambda_i\) their schemas and physical locators. Modalities keep independent index spaces, so lengths and correspondences are not forced, and absent modalities are declared rather than imputed. Episodes are published in immutable catalogs that record facts rather than pre-cut windows, enabling lazy materialization through byte-offset indices. Sampling is expressed by the model: the factory materializes exactly what the model requests without interpreting semantics, so a new source costs a description rather than a pipeline. 

\subsection{Quality Control and Gating}
\label{sec:quality-gating}

Quality control is implemented as a set of boundaries along the sample path, each with different granularity and cost. Offline gates operate on whole Episodes and windows, rejecting timelines with non-finite values or magnitudes beyond per-source guards, filtering stationary windows, and excluding Episodes shorter than a training window. Egocentric recordings receive an additional filter that removes windows dominated by camera self-motion, keeping only start positions whose rotation and translation stay within calibrated thresholds. Online gates at load time validate action and state ranges before video decoding, retry missing or corrupt videos against a bounded number of alternates, and blacklist persistent failures. Batch-level checks validate shapes, value ranges, finiteness, and the model input contract; a genuinely absent modality is marked rather than filled. At the step level, samples with extreme unified-action magnitudes have their loss weight zeroed, non-finite values are replaced before zeroing, and any step affected by non-finite parameters, loss, or gradients is skipped. The depth stream carries its own gates: a selection file marks usable Episodes, depth frames are checked against video frames for alignment, unreadable depth is marked absent and loses its loss share, and depths are clamped to the codec range.

\subsection{Data Preparation and Throughput Matching}
\label{sec:data-prep-throughput}

Reading and encoding a batch dominates preparation cost, while the noise used for training changes every step. We therefore encode each batch once and reuse the latents across \(N\) optimization steps, each of which draws fresh noise and timestep and runs a full forward--backward pass:
\begin{equation}
\label{eq:vae-reuse}
\mathcal{B}_t = \mathcal{B}_{t_0}, \qquad t_0 = \lfloor t / N \rfloor N,
\end{equation}
where \(\mathcal{B}_t\) is the encoded batch at step \(t\). The \(N\) steps are independent denoising objectives because the flow-matching target is rebuilt from fresh noise each step. This divides the encoding cost by \(N\).

\begin{figure}[!ht]
    \centering
    \includegraphics[width=\textwidth]{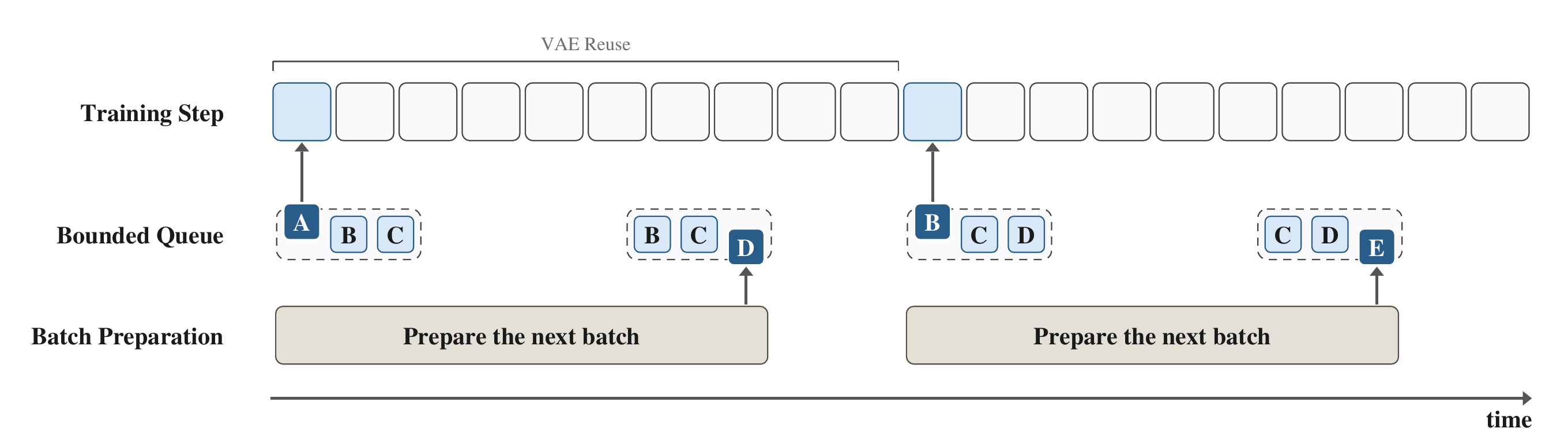}
    \caption{Throughput matching between data preparation and model computation. Batches are labelled in arrival order; preparation stays ahead of consumption within each VAE reuse window, so the model never waits.}
    \label{fig:throughput}
\end{figure}

To keep the training loop from waiting, a bounded per-rank queue lets a background producer keep complete batches ready ahead of consumption. Because a fresh batch is requested once every \(N\) steps, preparation stays hidden as long as
\begin{equation}
\label{eq:throughput-match}
C_{\text{prep}} \;\le\; N \cdot C_{\text{step}},
\end{equation}
where \(C_{\text{prep}}\) is host time to prepare a batch and \(C_{\text{step}}\) the time of one model step. When this holds with margin, the queue stays filled and the model never observes preparation latency; otherwise the queue drains and the model waits. Figure~\ref{fig:throughput} illustrates the arrangement.

\subsection{Load Balancing across Ranks}
\label{sec:load-balancing}

\begin{figure}[!ht]
    \centering
    \includegraphics[width=0.7\textwidth]{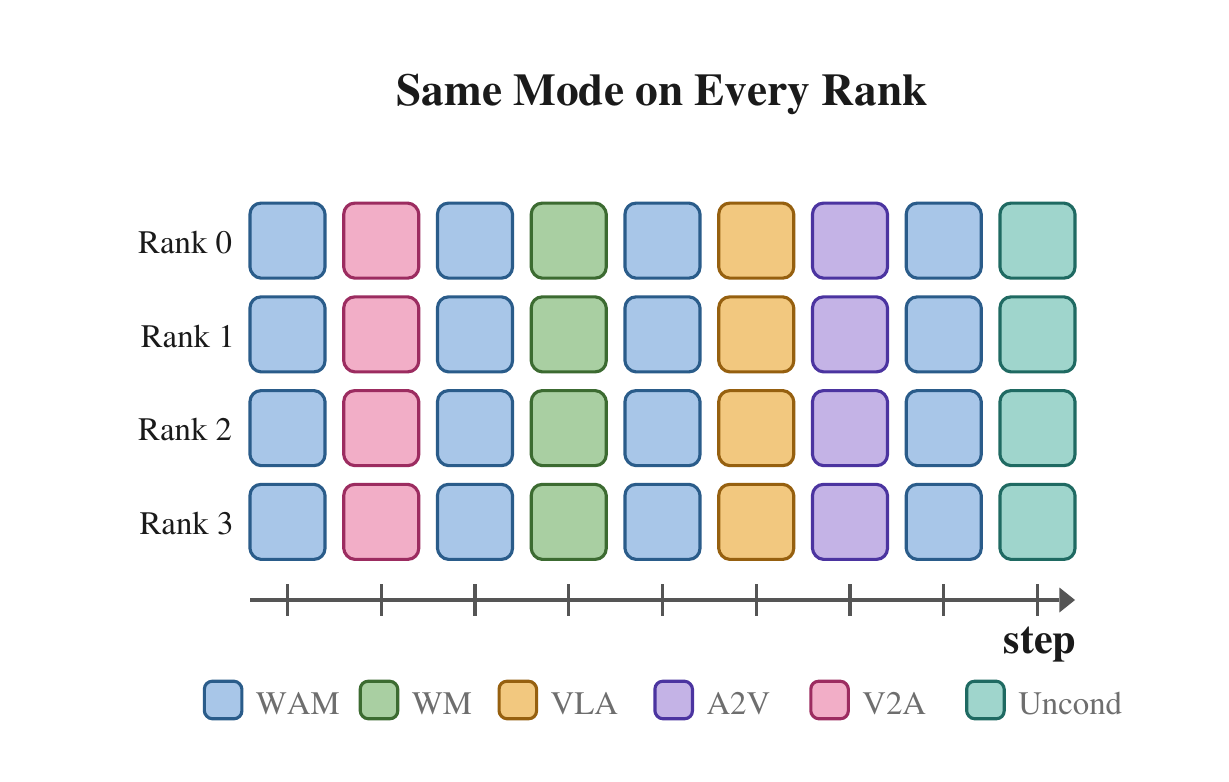}
    \caption{Load balancing across ranks in a synchronous run. Every row is one
    rank and every column one optimization step, and the function mode of a step
    is sampled with a generator seeded only by the step number, so a column
    carries the same mode on every rank and the same choice is reproduced after
    a restart. The legend names the six modes, which are sampled in the
    configured proportions.
}
    \label{fig:load-balancing}
\end{figure}

In synchronous training, all ranks must update the same parameters at the same step. If ranks train different branch subsets, the reduction operates over different tensor sets and can deadlock under ZeRO-1. We therefore make the function mode a pure function of the step number, sampled from configured proportions with a generator seeded only by that step, persisted with the checkpoint, and advanced identically by every rank. This guarantees agreement across ranks and across restarts. The same reasoning requires identical token layouts: windows are cut to fixed lengths so no padding is needed and every rank carries the same number of tokens. Figure~\ref{fig:load-balancing} illustrates the arrangement.

Source assignment offers two deterministic schedules. The default assigns the same source to every rank at a given step, keeping batches internally consistent. The alternative distributes rank slots across sources in proportion to their weights, mixing the corpus more finely at the risk of stragglers. Both require no inter-rank communication. Losses are averaged across ranks, and non-finite and outlier signals are reduced so that a condition seen by one rank is acted on by all.

\subsection{Evaluation Orchestration}
\label{sec:eval-orchestration}

Evaluation runs every fixed number of steps on a fixed set of samples, with deterministic assignment to ranks derived from the same weights as the training mixture. To avoid competing with training for GPUs, a background process decodes evaluation clips into a shared-storage cache and writes a completion manifest. The foreground evaluation only reads the cache, so prepared evaluations take seconds and never occupy a GPU needed by training. The cache is content-keyed and reused across restarts and runs; an evaluation whose cache is not ready is skipped rather than stalling training. Out-of-memory shards are rerun with smaller micro-batches, keeping the evaluated set complete, and results join the same logging channel as training metrics.

\vspace{2cm}
\section{Training Strategy}
\label{sec:training-strategy}

Thinker-WM supports multiple control modes and is trained on heterogeneous data sources. Large-scale general video supplies broad visual and temporal priors~\cite{worldmodels2018,genie,pandora}; real robot recordings provide embodiment-specific visual statistics and the action supervision on which the conditionals of Eq.~\eqref{eq:joint} depend; egocentric human video supplies scalable task, scene, and object diversity that robot teleoperation cannot economically reach~\cite{rt1,oxe}. The Data Factory provides the execution substrate, and this section specifies how these sources are combined into a staged curriculum.

\subsection{Staged Training Curriculum}
\label{sec:curriculum}

Training across all data distributions from the outset can destabilize optimization, as heterogeneous objectives compete for the capacity of the shared model and may pull the representation toward dominant training distributions. Fully sequential training, in contrast, risks catastrophic forgetting of capabilities acquired in earlier stages~\cite{mccloskey1989catastrophic,kirkpatrick2017ewc}. To balance acquisition and retention, Thinker-WM adopts a staged curriculum with lightweight rehearsal, progressively introducing new data distributions and capabilities while maintaining a small replay stream from earlier stages~\cite{robins1995catastrophic,bengio2009curriculum}. The training distribution of stage \(s\) is
\begin{equation}
\label{eq:stage-mix}
\mathcal{D}_s \;=\; (1-\epsilon_s)\,\mathcal{D}_s^{\mathrm{main}} \;+\;
\epsilon_s\,\mathcal{D}_s^{\mathrm{replay}},
\qquad \epsilon_s \ll 1,
\end{equation}
where \(\mathcal{D}_s^{\mathrm{main}}\) carries the new capability and \(\mathcal{D}_s^{\mathrm{replay}}\) samples all previously aligned distributions. Each stage therefore mixes in a small fraction of the data from the preceding stage, so that capabilities accumulate rather than overwrite. Stages are ordered so that shared representations stabilize before downstream modules couple into them. Table~\ref{tab:training-stages} summarizes the data, updated scope, and installed capability of each stage.

\begin{table}[ht]
  \centering
  \caption{The four training stages. Each stage installs one or more capabilities while mixing in a fixed small fraction of the data from the previous stage. Stage~0 is the pretrained Wan2.2 model, used as it stands.}
  \label{tab:training-stages}
  \small
  \hyphenpenalty=10000\exhyphenpenalty=10000
  \begin{tabular}{@{}c p{4.7cm} l l@{}}
    \toprule
    \textbf{Stage} & \textbf{Main data} & \textbf{Updated scope} & \textbf{Capability installed}\\
    \midrule
    0 & internet-scale video & inherited & visual and world prior\\
    \midrule
    1 & robot data\newline
        egocentric human video\newline
        data from the previous stage
      & video pathway & robot-view video prediction\\
    \midrule
    2 & curated robot data\newline
        curated egocentric human video\newline
        data from the previous stage
      & all parameters & video-action coupling + geometry\\
    \midrule
    3 & target-robot data\newline
        data from the previous stage
      & all parameters & target-robot specialization\\
    \bottomrule
  \end{tabular}
\end{table}

Stage~0 is the pretrained Wan2.2 TI2V-5B backbone~\cite{wan}, which supplies the broad visual and temporal prior and is not trained here. Stage~1 introduces open-source robot recordings and egocentric human video, training only the video pathway to adapt the inherited prior to first-person embodied observations; a small fraction of Stage~0 data is replayed to preserve scene, lighting, and camera-motion diversity. Stage~2 turns the video-only model into a joint video--action model on curated robot and egocentric data, introducing the action stream and the geometric stream together, with the planning condition providing a shared conditioning signal across branches. Stage~3 specializes the model to the target robot using embodiment-specific recordings and accurate geometric supervision; the earlier stages are replayed so that the shared backbone and previously acquired capabilities are preserved. Overall, the curriculum follows a progression of \textit{inheritance}, \textit{integration}, and \textit{specialization}.

\subsection{Stage-wise Parameter and Geometry Schedule}
\label{sec:parameter-geometry}

We initialize from the Wan2.2 TI2V-5B backbone~\cite{wan} and reuse its pretrained blocks, patch embeddings, modulation layers, and positional geometry across all three architectural depths (Section~\ref{sec:backbone-split}). Stage~0 leaves these weights untouched. Stage~1 updates the visual pathway only, adapting the inherited prior to robot-view and first-person statistics without disturbing branches that have no counterpart in the pretrained model. From Stage~2 onward, all trainable parameters are active: the action tower, the geometric pathway, and the shared UDT are optimized jointly. The frozen components remain fixed throughout, as the VAE and the VLM keep their pretrained weights.

Geometry enters in Stage~2 and is integrated into the shared representation so that the model jointly predicts future observations, geometry, and actions. While the mixture is dominated by data with coarse or partial depth annotation, the geometric channel is one-way: geometry reads appearance and action, but neither reads geometry, so the geometric objective shapes the shared weights only through its own loss gradient. Once training is restricted to datasets with accurate depth, which in this curriculum is the target-robot fine-tuning stage, the remaining direction is opened so that appearance and action also read the geometric stream (Figures~\ref{fig:attention-flow} and~\ref{fig:attention-flow-sft}).

\subsection{Function Mode and Noise Scheduling}
\label{sec:mode-noise}

The function modes of Section~\ref{sec:function-training} partition the joint distribution of Eq.~\eqref{eq:joint} across world modeling, policy learning, and cross-modal generation. They are sampled in the fixed ratio
\[
\texttt{wam}:\texttt{wm}:\texttt{vla}:\texttt{a2v}:\texttt{v2a}:\texttt{uncond}
= 0.5:0.1:0.1:0.1:0.1:0.1,
\]
so that half of the optimization steps train the full joint distribution while every function is exercised regularly. Mode assignment is made once per optimization step for the whole batch, seeded by the global step index and identical across all data-parallel ranks, so every rank computes gradients for the same set of branches and the sharded optimizer remains consistent. These proportions belong to the training configuration and can be revised without touching the architecture.

Both streams are trained against the time-shifted noise grids of Eq.~\eqref{eq:shift}. The shift concentrates video training at high noise levels, where global scene structure is determined and the 3900-token visual latent requires the most supervision, while the 48-token action sequence is low-dimensional and smooth, so a near-uniform grid covers it without wasting steps. Within the joint modes \texttt{wam} and \texttt{uncond}, the two noise levels share a common base \(u\) and receive independent jitter \(\delta_m\) before applying their own shift:
\begin{equation}
\label{eq:jitter}
\tilde{\sigma}_m \;=\; \mathrm{clamp}\bigl(u + \delta_m,\, 0,\, 1\bigr),
\qquad
u \sim \mathrm{Uniform}(0,1),
\qquad
\delta_m \sim \mathrm{Uniform}(-r, r),
\qquad
m \in \{v, a\}.
\end{equation}
At inference the two streams are denoised in lockstep along coupled trajectories, so the noise combinations visited during training should match those traversed at inference. The shared base keeps the streams at matched denoising progress, and the jitter half-width \(r=0.10\), calibrated to one cell of the 10-step action inference grid, decorrelates their exact levels within that resolution. Single-modality modes retain independent per-modality sampling, since only one stream carries a loss there.



\vspace{2cm}
\section{Motion Probe}
\label{sec:probe}

Chapters~\ref{sec:model-architecture} and~\ref{sec:training-strategy} establish that Thinker-WM jointly models RGB, depth, and action through a shared Unified Dynamics Transformer (UDT). A natural question is whether the shared representation truly captures cross-modal coupling, or merely processes modalities in parallel. We introduce \textbf{Motion Probe}, a lightweight diagnostic module that evaluates whether motion information can be recovered from the shared video representation, thereby providing direct evidence for the effectiveness of the UDT. Motion probe's effect is evaluated in Section~\ref{sec:probe-eval}.
\begin{figure}[ht]
    \centering
    \includegraphics[width=\linewidth]{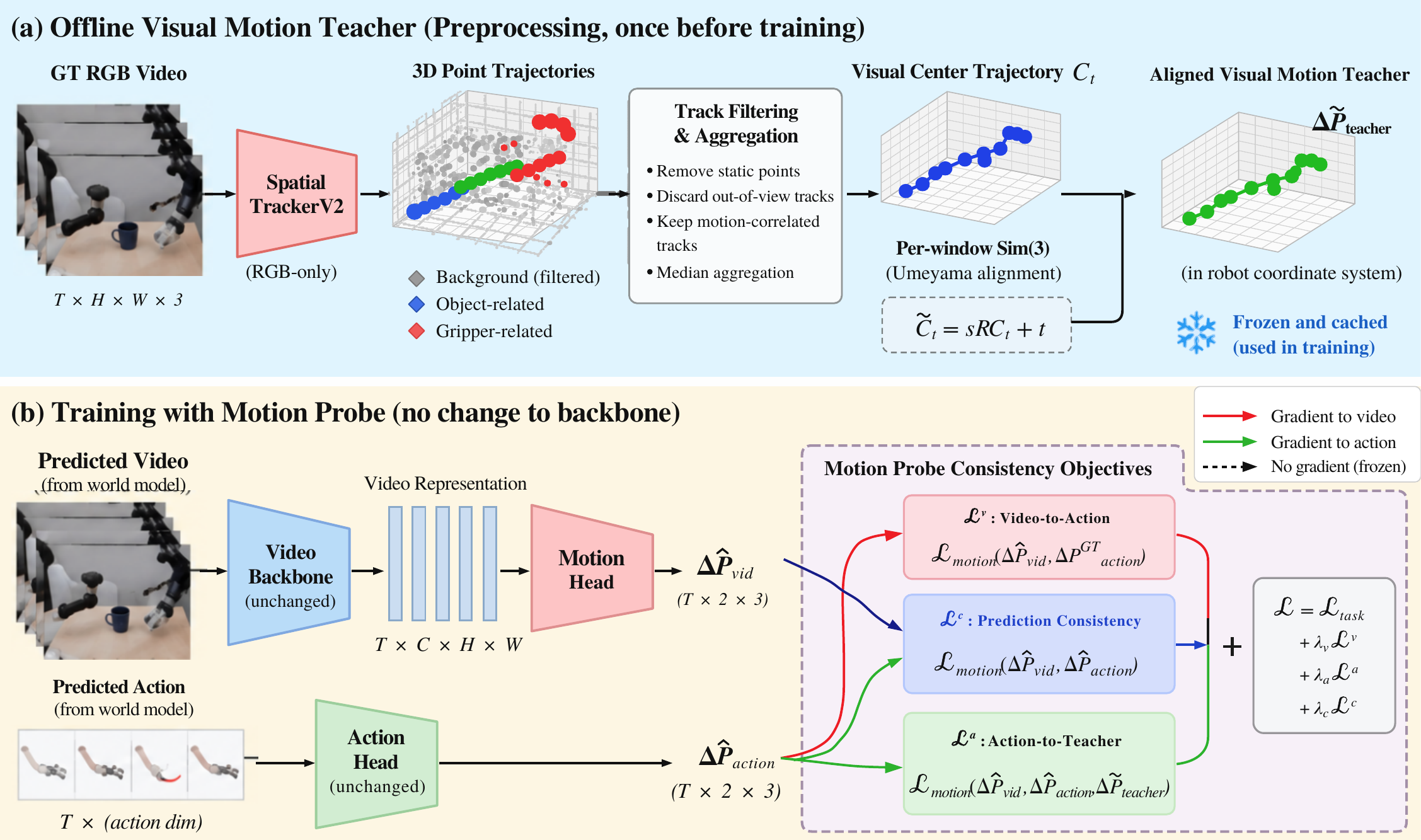}
    \caption{
    Overview of Motion Probe.
    \textbf{(a)} Offline visual motion teacher construction: SpatialTrackerV2 extracts 3D point trajectories from ground-truth RGB video, motion-aware filtering and median aggregation produce a robust visual center trajectory, and a per-window Sim(3) transform aligns it with the recorded end-effector motion. The aligned teacher is frozen and cached.
    \textbf{(b)} Probing the frozen UDT: a lightweight Motion Head predicts end-effector displacement from the shared video representation. The UDT weights remain frozen, and only the Motion Head is trained. Low probe error indicates that the shared representation encodes action-relevant motion.
    }
    \label{fig:probe-agibot-pipeline}
\end{figure}
\subsection{Probe Design and Offline Motion Teacher}
\label{sec:probe-teacher}

The probe rests on a simple logic: if the UDT truly unifies appearance and action, the video tokens should carry sufficient motion information to reconstruct end-effector displacement aligned with the executed action, even without action-specific supervision on the video pathway. We test this by attaching a lightweight \textbf{Motion Head} to the frozen UDT video tokens. The head is deliberately small---spatial pooling followed by temporal regression---so that its capacity cannot compensate for missing motion information in the shared representation. The UDT weights are kept frozen throughout the probe, and only the Motion Head is trained.

To evaluate the probe on real-robot data, we require a reliable visual motion reference. However, AgiBot~\cite{agibotworld} lacks reliable camera extrinsics and depth. We therefore construct an offline motion teacher, in which SpatialTrackerV2~\cite{xiao2025spatialtrackerv2} is applied once to the ground-truth RGB video to produce 3D point trajectories on a dense grid. We discard tracks with negligible image-space motion or that leave the valid image region for a large fraction of the sequence, retain tracks whose motion magnitude correlates with the recorded end-effector motion, and aggregate them by temporal median to obtain a robust visual center trajectory \(C_t\). An Umeyama alignment between \(C_t\) and the ground-truth end-effector trajectory yields a per-window Sim(3) transform parameterized by scale \(s\), rotation \(R\), and translation \(t\):
\begin{equation}
\tilde{C}_t = s R C_t + t.
\end{equation}
Since the probe is defined on temporal displacement, the translation cancels after differencing:
\begin{equation}
\Delta \tilde{C}_t
=
\tilde{C}_{t+1}-\tilde{C}_t
=
sR\left(C_{t+1}-C_t\right)
=
sR\Delta C_t.
\end{equation}
The aligned displacement \(\Delta \tilde{C}_t\) is the visual motion teacher. The Sim(3) parameters are estimated once from ground-truth data and frozen; they are never refined on model predictions, so the teacher remains a fixed geometric reference rather than an adaptive target.

Not every window yields reliable supervision. A window is marked \textbf{usable} if path correlation \(\ge 0.95\), motion-magnitude correlation \(\ge 0.55\), trajectory-shape agreement \(\ge 0.65\), frame-wise directional cosine \(\ge 0.5\), and the estimated scale \(s \in [0.2, 30]\). Windows that fail these criteria are excluded from the probe. This filtering is applied offline, and the resulting teacher signals are cached.

\subsection{Probe Objective and Interpretation}
\label{sec:probe-objective}

The Motion Head maps frozen UDT video tokens to left- and right-hand translational displacements,
\begin{equation}
\Delta \hat{P}_{\mathrm{vid}}
\in
\mathbb{R}^{T \times 2 \times 3}.
\end{equation}
It is trained only on the motion loss between predicted video motion and ground-truth end-effector displacement:
\begin{equation}
\mathcal{L}_{\mathrm{probe}}
=
\mathcal{L}_{\mathrm{motion}}
\left(
\Delta \hat{P}_{\mathrm{vid}},
\Delta P_{\mathrm{action}}^{\mathrm{GT}}
\right),
\end{equation}
where \(\mathcal{L}_{\mathrm{motion}}\) combines end-point error and motion-magnitude discrepancy:
\begin{equation}
\mathcal{L}_{\mathrm{motion}}(x,y)
=
\frac{1}{|\mathcal{T}|}
\sum_{t \in \mathcal{T}}
\|x_t-y_t\|_2
+
\frac{1}{|\mathcal{T}|}
\sum_{t \in \mathcal{T}}
\left|
\|x_t\|_2-\|y_t\|_2
\right|.
\end{equation}

The probe is interpreted as a measurement, not as a performance improvement. If the frozen UDT video representation already encodes action-relevant motion, a small head should recover displacement with low error; if recovery fails, the shared representation does not carry sufficient motion structure for downstream use. A low probe error therefore provides direct evidence that the UDT has learned a genuinely unified representation in which appearance and action are coupled, rather than merely processed in parallel. The probe introduces no new training stage, no modification to the backbone, and no additional generation branch; its value lies in the representational evidence it provides.

\vspace{2cm}
\section{Experiments}
\label{sec:experiment}
\subsection{Results on Benchmarks}
We report our evaluation results on three well-recognized world-model benchmarks.
\subsubsection{EWMBench Evaluation}
\label{sec:experiment-ewmbench}

We first evaluate the quality of the generated world itself. EWMBench~\cite{ewmbench}
is a standard benchmark for embodied world models. It scores a generated video
along eight dimensions covering visual scene consistency, end-effector motion
correctness, and semantic alignment with the instruction, and reports their sum
as Overall. Its standard evaluation set contains 21 episodes drawn from 7
manipulation tasks. For every episode, the model produces three candidate videos
conditioned on the ground-truth first frame and the official instruction, the best
candidate is selected by Hausdorff distance, and each dimension is normalized to $[0,1]$ and averaged over all valid trials, 63 videos in total. Generation uses
48 frames at a resolution of $480\times640$ with 50 denoising steps and a
classifier-free guidance scale of 5.0. Since the benchmark computes its metrics
at 30 FPS, generated results are upsampled from 5 FPS to 30 FPS so that the
comparison follows the protocol on which the benchmark is defined.

\begin{table}[ht]
  \centering
  \caption{Generation quality on EWMBench. Every dimension is normalized to
  $[0,1]$, the Overall column is the sum of the eight dimensions, and higher is
  better. Numbers for the baseline models are taken from
  Qwen-RobotWorld~\cite{qwenrobotworld}, which reports them on the same
  evaluation set. The last block is our model.}
  \label{tab:experiment-ewmbench}
  \small
  \resizebox{\textwidth}{!}{%
  \begin{tabular}{@{}llccccccccc@{}}
    \toprule
    \textbf{Type} & \textbf{Model} & \textbf{SceneC} $\uparrow$ & \textbf{HSD} $\uparrow$ & \textbf{Dyn} $\uparrow$ & \textbf{nDTW} $\uparrow$ & \textbf{Diversity} $\uparrow$ & \textbf{BLEU} $\uparrow$ & \textbf{CLIP} $\uparrow$ & \textbf{Logics} $\uparrow$ & \textbf{Overall} $\uparrow$\\
    \midrule
    General video & Veo3 & 0.8415 & 0.2130 & 0.1932 & 0.1613 & 0.0221 & 0.2139 & 0.8965 & 0.9474 & 3.49\\
     & Wan2.6 & 0.6712 & 0.2034 & 0.0900 & 0.1715 & 0.0502 & 0.1616 & 0.8743 & 1.0000 & 3.22\\
     & Kling26 & 0.8211 & 0.3272 & 0.1822 & 0.3423 & 0.0173 & 0.2591 & 0.9014 & 1.0000 & 3.85\\
     & LTX-2 & 0.7850 & 0.2076 & 0.1283 & 0.2443 & 0.0120 & 0.1425 & 0.8869 & 0.5000 & 3.01\\
     & Sora2 & 0.8526 & 0.2807 & 0.3494 & 0.2754 & 0.0314 & 0.2466 & 0.9100 & 0.9474 & 3.89\\
    \midrule
    Embodied world model & Cosmos & 0.7963 & 0.2500 & 0.2052 & 0.2533 & 0.0803 & 0.1230 & 0.8458 & 0.7333 & 3.29\\
     & GigaWorld & 0.8707 & 0.3050 & 0.0849 & 0.2783 & 0.0278 & 0.2048 & 0.8873 & 0.9000 & 3.56\\
     & LVP & 0.8795 & 0.4248 & 0.0433 & 0.6226 & 0.0093 & 0.2179 & 0.8995 & 0.9524 & 4.05\\
     & Vidar & 0.7341 & 0.1877 & 0.1520 & 0.1769 & 0.0653 & 0.1607 & 0.8821 & 0.9411 & 3.30\\
     & WoW & 0.8866 & 0.2494 & 0.0529 & 0.2566 & 0.0266 & 0.1932 & 0.9001 & 0.9524 & 3.52\\
     & Qwen-RobotWorld & 0.9142 & 0.5660 & 0.3429 & 0.6708 & 0.0114 & 0.2079 & 0.8834 & 1.0000 & 4.60\\
    \midrule
    \textbf{Ours} & \textbf{Thinker-WM} & \textbf{0.8877} & \textbf{0.5638} & \textbf{0.3630} & \textbf{0.7102} & \textbf{0.0117} & \textbf{0.2691} & \textbf{0.9020} & \textbf{0.9524} & \textbf{4.66}\\
    \bottomrule
  \end{tabular}%
  }
\end{table}

Table~\ref{tab:experiment-ewmbench} reports the main results. General video
generators reach high visual scores but degrade sharply on the motion
dimensions. Veo3 scores 0.8415 on SceneC and 0.8965 on CLIP, yet its HSD and
Dyn fall to 0.2130 and 0.1932, showing that visual fidelity alone does not
imply a physically consistent world. Embodied world models close part of this
gap, and Thinker-WM obtains the highest Overall in the table at 4.66. The
advantage is most visible where appearance and action meet: nDTW reaches
0.7102 against 0.6708 for Qwen-RobotWorld and 0.6226 for LVP, and Dyn reaches
0.3630 against 0.3429 and 0.0433. This is consistent with the model design:
video, depth, and action are generated in one joint process rather than through
an action head attached to a video generator, so the robot trajectory is part
of what the generative process is trained to match.

\begin{figure}[ht]
  \centering
  \includegraphics[width=0.9\textwidth]{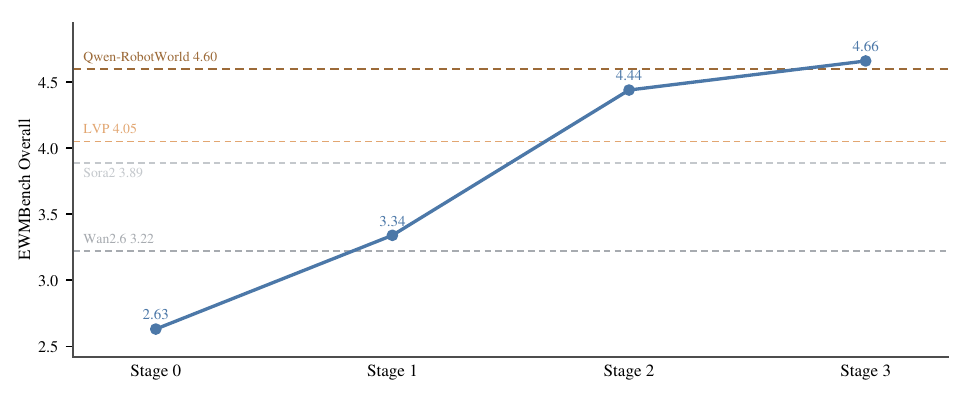}
  \caption{Overall EWMBench score across the four training stages of
  Section~\ref{sec:training-strategy}, with the final checkpoint of each
  stage as a point. The dashed lines are four reference baselines from
  Table~\ref{tab:experiment-ewmbench}, namely Wan2.6, Sora2 as the strongest
  general video model, LVP, and the strongest overall, Qwen-RobotWorld.}
  \label{fig:experiment-stage-curve}
\end{figure}

The same benchmark also tracks the four stages of the training curriculum in
Section~\ref{sec:training-strategy}. The pretrained backbone, used without any
training of our own, obtains an Overall of 2.63. Training the video pathway
alone on robot first-person data raises it to 3.34, joint video--action
training reaches 4.44, and supervised fine-tuning on the recordings of the
deployed robot reaches 4.66. The progression is monotone, so both the breadth
of the mixture and the specialization to the deployed robot translate into
measurable generation quality (Figure~\ref{fig:experiment-stage-curve}).

\subsubsection{WorldOlympiad Evaluation}
\label{sec:experiment-worldolympiad}

WorldOlympiad~\cite{2026WorldOlympiad} complements EWMBench by evaluating video
world models on a broader real-robot distribution. We use its Robotics
subdomain leaderboard, which contains 400 manipulation episodes from more than ten
physical robot embodiments and spans multiple camera configurations, including
head-mounted, wrist-mounted, and external viewpoints. The benchmark reports
physical faithfulness, 3D consistency, CLIP-based semantic similarity, and
interaction fidelity, providing a stronger test of cross-embodiment and
cross-view generalization than EWMBench. Both benchmarks are built on real-robot
interaction videos, so together they assess whether the learned visual dynamics
remain consistent across heterogeneous real-world robotic settings.

Following the official WorldOlympiad protocol, the
overall score, denoted as \textbf{All} in Table~\ref{tab:worldolympiad_robotics}, is computed
as the equal-weight average of the three core evaluation tracks:
\begin{equation}
    S_{\mathrm{All}}
    =
    \frac{1}{3}
    \left(
        S_{\mathrm{phys}}
        +
        S_{\mathrm{3D}}
        +
        S_{\mathrm{interact}}
    \right),
\end{equation}
where $S_{\mathrm{phys}}$, $S_{\mathrm{3D}}$, and $S_{\mathrm{interact}}$
denote physical faithfulness, 3D consistency, and interaction fidelity,
respectively. The reported CLIP score is not averaged as an additional
independent term in $S_{\mathrm{All}}$; rather, it serves as an auxiliary
semantic signal within the interaction metric.

\begin{table}[htbp]
    \centering
    \caption{WorldOlympiad Evaluation Results on the Robotics Subdomain Leaderboard.}
    \label{tab:worldolympiad_robotics}
    \begin{tabular}{lccccc}
        \toprule
        \textbf{Model} &
        \textbf{Physical} &
        \textbf{3D Cons.} &
        \textbf{CLIP} &
        \textbf{Interact.} &
        \textbf{All} \\
        \midrule
        Uranus
        & 0.925
        & 0.511
        & 0.304
        & 0.730
        & 0.722 \\
        Cosmos-Predict-2.5
        & 0.937
        & 0.479
        & 0.321
        & 0.721
        & 0.712 \\
        LingBot-World
        & \textbf{0.949}
        & 0.393
        & 0.314
        & 0.710
        & 0.684 \\
        Rolling Forcing
        & 0.870
        & 0.389
        & \textbf{0.329}
        & 0.566
        & 0.608 \\
        LongLive
        & 0.857
        & 0.472
        & 0.327
        & 0.470
        & 0.600 \\
        Yume-1.5
        & 0.851
        & 0.288
        & 0.312
        & 0.624
        & 0.588 \\
        WoW
        & 0.787
        & 0.272
        & 0.288
        & 0.447
        & 0.502 \\
        \midrule
        \textbf{Ours}
        & 0.923
        & \textbf{0.531}
        & 0.306
        & \textbf{0.765}
        & \textbf{0.740} \\
        \bottomrule
    \end{tabular}
\end{table}

\begin{figure}[htbp]
    \centering
    \includegraphics[width=0.9\textwidth]{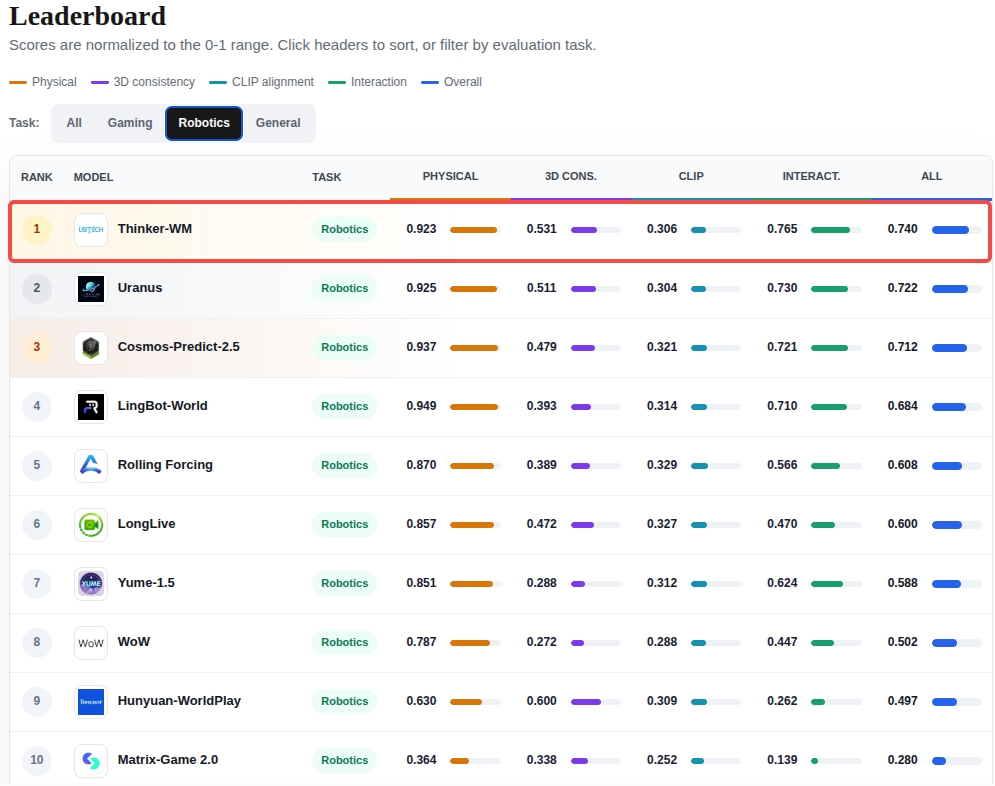}
    \caption{Official screenshot of the WorldOlympiad robotics leaderboard, where our proposed model ranks first.}
    \label{fig:worldolympiad_leaderboard}
\end{figure}

As shown in Table~\ref{tab:worldolympiad_robotics}, our proposed Thinker-WM achieves the
highest overall score of 0.740, together with the best 3D Consistency (0.531)
and Interaction score (0.765) among the compared methods. These results
indicate a favorable balance between geometric consistency and interaction
fidelity, supporting the model's ability to produce plausible and coherent
future dynamics in real-world robotic scenarios.
Figure~\ref{fig:worldolympiad_leaderboard} presents the official screenshot of the WorldOlympiad robotics leaderboard, corroborating our top-ranking performance on this benchmark.

\subsubsection{WorldArena Track-1 Evaluation}
\label{sec:experiment-worldarena-track1}


We further evaluate on WorldArena~\cite{2026WorldArena} Track-1, a
simulation-based benchmark for open-loop embodied video generation. Unlike
EWMBench and WorldOlympiad, which are built on real-robot manipulation videos,
WorldArena provides controlled simulated environments with diverse task
configurations and interaction dynamics. The track assesses visual appearance,
temporal consistency, physical plausibility, 3D geometric fidelity, and
compliance with task instructions and action conditions. Because Dynamic
Degree, Flow Score, and Motion Smoothness are highly sensitive to frame rate
and temporal sampling density, we exclude these three motion-related metrics
and adopt the unweighted mean of the remaining twelve metrics as the core
comparison metric, following the official benchmark recommendation as these three motion metrics suffer from frame-sampling exploitation bugs. The retained metrics cover visual quality, content
consistency, physics adherence, 3D accuracy, and controllability: \textit{Image
Quality}, \textit{Aesthetic Quality}, \textit{JEPA Similarity}, \textit{Subject Consistency}, \textit{Background Consistency}, \textit{Photometric Consistency}, \textit{Interaction Quality, Trajectory}, \textit{Accuracy}, \textit{Depth Accuracy}, \textit{Perspectivity}, \textit{Instruction Following}, and \textit{Semantic Alignment}.

\begin{table}[ht]
\centering
\caption{
WorldArena dimension-level scores under the 12-metric protocol.
}
\label{tab:worldarena_12_metrics}
\resizebox{\textwidth}{!}{%
\begin{tabular}{llccccccc}
\toprule
& \textbf{Metric}
& \textbf{Ours}
& \textbf{DexWorldEngine}
& \textbf{BWM-Fast}
& \textbf{UNIS}
& \textbf{MotuBrain}
& \textbf{Wan2.6}
& \textbf{Cosmos-Predict2.5} \\
\midrule

\multirow{3}{*}{Visual Quality}
& Image Quality
& 37.26
& 51.44
& 51.22
& 53.94
& 54.05
& \textbf{67.36}
& 49.48 \\

& Aesthetic Quality
& 37.65
& 40.96
& 40.15
& 40.79
& 41.02
& \textbf{44.40}
& 34.14 \\

& JEPA Similarity
& 93.94
& 86.71
& \textbf{97.87}
& 90.60
& 96.88
& 72.57
& 90.65 \\
\midrule

\multirow{3}{*}{Content Consistency}
& Subject Consistency
& 82.58
& 81.46
& 81.42
& 79.05
& \textbf{83.51}
& 73.15
& 67.24 \\

& Background Consistency
& \textbf{90.19}
& 88.85
& 90.17
& 86.44
& 89.91
& 84.29
& 72.72 \\

& Photometric Consistency
& \textbf{46.33}
& 4.42
& 3.08
& 2.13
& 5.10
& 34.57
& 45.21 \\
\midrule

\multirow{2}{*}{Physics Adherence}
& Interaction Quality
& 70.20
& 81.04
& 79.88
& 87.30
& 73.90
& 73.16
& \textbf{90.35} \\

& Trajectory Accuracy
& 50.71
& \textbf{51.41}
& 44.89
& 41.89
& 43.10
& 12.18
& 27.49 \\
\midrule

\multirow{2}{*}{3D Accuracy}
& Depth Accuracy
& \textbf{92.49}
& 86.19
& 86.41
& 85.25
& 88.65
& 75.43
& 60.54 \\

& Perspectivity
& 90.08
& 97.58
& 97.14
& \textbf{98.84}
& 94.32
& 93.94
& 84.68 \\
\midrule

\multirow{2}{*}{Controllability}
& Instruction Following
& 78.86
& 92.62
& 90.22
& 93.86
& 82.60
& \textbf{93.94}
& 90.35 \\

& Semantic Alignment
& \textbf{90.47}
& 89.45
& 89.15
& 89.35
& 89.09
& 88.09
& 89.86 \\
\midrule

\multicolumn{2}{l}{\textbf{12-Metric EWMScore} $\uparrow$}
& \textbf{71.73}
& 71.01
& 70.97
& 70.79
& 70.18
& 67.43
& 64.40 \\
\bottomrule
\end{tabular}%
}
\end{table}

Table~\ref{tab:worldarena_12_metrics} reports the 12-metric evaluation. Our
model achieves the highest overall score of 71.73, with particularly strong
performance in Background Consistency, Photometric Consistency, Depth Accuracy,
and Semantic Alignment. Together with the real-world evaluations on EWMBench
and WorldOlympiad, the WorldArena results provide complementary evidence that
the video prediction component of Thinker-WM maintains strong visual and
structural prediction capability across both real and simulated embodied
environments.

\subsection{Ablation and Analysis}
We conduct a set of controlled experiments to examine the functional behavior of
Thinker-WM and the role of its shared multimodal representation. The experiments
address four complementary questions. First, does the model actually consume an
action sequence when it is provided as a condition? Second, does depth contribute
to downstream control beyond improving geometric prediction, and does the direction
of RGB--depth interaction matter? Third, can the six function modes share the same
UDT parameters without introducing substantial optimization conflict? Finally, can
additional motion-level supervision improve action prediction without degrading the
existing video representation?

\subsubsection{Action Conditioning Controls Future Generation}
\label{sec:experiment-joint-video-action}

A central capability of a world--action model is to let a specified action trajectory
determine the predicted future rather than merely coexisting with an independently
generated video. We therefore compare the standard joint-generation mode with an
action-conditioned mode in which the ground-truth action sequence is provided as a
clean condition and the language prompt is removed. Both settings use the same
network parameters, so the comparison isolates the effect of conditioning the video
generation process on an explicit action trajectory.

\begin{table}[ht]
  \centering
  \caption{Generation with the ground-truth action sequence as a condition,
  compared with joint generation on the same model. Numbers are averaged over
  the evaluated checkpoints, and the dimensions that compare motion against the
  ground truth are shown separately.}
  \label{tab:experiment-action-conditioned}
  \small
  \begin{tabular}{@{}lccc@{}}
    \toprule
    \textbf{Dimension} & \textbf{Joint} & \textbf{Action-conditioned} & \textbf{$\Delta$}\\
    \midrule
    Overall & 4.66 & 5.03 & $+0.37$\\
    HSD & 0.5638 & 0.7369 & $+0.1731$\\
    nDTW & 0.7102 & 0.8736 & $+0.1634$\\
    Dyn & 0.3630 & 0.3636 & $+0.0006$\\
    \bottomrule
  \end{tabular}
\end{table}

Table~\ref{tab:experiment-action-conditioned} reports the results averaged over the
evaluated checkpoints. Supplying the ground-truth action increases the Overall score
from $4.66$ to $5.03$, with the largest improvements appearing in the motion-sensitive
metrics. HSD increases from $0.5638$ to $0.7369$, nDTW from $0.7102$ to $0.8736$, and
Dyn from $0.3630$ to $0.3636$. In particular, HSD and nDTW directly measure the
agreement between the generated motion and the recorded trajectory, making their
improvement a direct indication that the predicted future changes in response to the
provided action sequence.

The result shows that the action stream is used as a generative condition rather than
treated as an auxiliary output that can be ignored by the video pathway. The same
weights can therefore represent both autonomous future generation and action-steered
future prediction, with the supplied trajectory providing an explicit control signal for the generated world. Qualitative trajectory-transfer results in
Section~\ref{sec:experiment-trajectory-transfer} provide a complementary visualization of this behavior.

\subsubsection{Depth as a Co-Evolving Generative State}
\label{sec:experiment-depth}

We evaluate depth from two complementary perspectives: the accuracy of
generated geometry and the benefit of RGB--depth interaction for
closed-loop control.

\paragraph{Depth generation accuracy.}
We compare Thinker-WM with RynnWorld-4D~\cite{zhao2026rynnworld} on 48 AgiBot
validation sequences. RynnWorld-4D also includes AgiBot in its training
corpus. We evaluate generated depth using absolute relative error
(AbsRel) and root mean squared error (RMSE).
As shown in Table~\ref{tab:depth_generation}, Thinker-WM achieves an
AbsRel of $0.1360$ and an RMSE of $0.1790$\,m, compared with $0.1625$
and $0.2075$\,m for RynnWorld-4D. These correspond to relative error
reductions of $16.3\%$ and $13.7\%$, respectively.

As a valid-range sensitivity check, restricting RynnWorld-4D's
evaluation range from $0.05$--$5.0$\,m to the $0.15$--$4.0$\,m range
used by Thinker-WM changes its AbsRel only from $0.162517$ to
$0.162474$. Thus, the observed AbsRel improvement is not explained
by this difference in valid depth range.

\begin{table}[ht]
    \centering
    \caption{
    Generated depth accuracy on 48 AgiBot validation sequences.
    RMSE is measured in meters. Lower is better for both metrics.
    }
    \label{tab:depth_generation}
    \small
    \begin{tabular}{lcc}
        \toprule
        \textbf{Method} &
        \textbf{AbsRel $\downarrow$} &
        \textbf{RMSE (m) $\downarrow$} \\
        \midrule
        RynnWorld-4D
        & 0.1625
        & 0.2075 \\

        \textbf{Thinker-WM}
        & \textbf{0.1360}
        & \textbf{0.1790} \\
        \bottomrule
    \end{tabular}
\end{table}

\paragraph{Impact on closed-loop control.}
Accurate depth generation alone does not establish its utility for
action prediction. We therefore conduct a controlled ablation on the
RoboTwin \texttt{turn\_switch} task. All three configurations are trained
for 8K steps under the same training recipe, using 24 video frames
and a 48-step action horizon. At inference, video and action are each
sampled with 10 denoising steps. We evaluate the 6K, 7K, and 8K
checkpoints with 20 closed-loop trials per checkpoint. The \textbf{RGB-only} baseline removes the depth stream.
\textbf{RGB + Depth (One-way)} adds depth generation but restricts
RGB--depth interaction to one direction.
Our full \textbf{Thinker-WM} enables bidirectional RGB--depth
interaction within the shared generative backbone, allowing appearance
and geometry to co-evolve before decoding. This comparison tests
whether bidirectional interaction provides benefits beyond introducing
a depth stream with restricted coupling.

\begin{table}[ht]
    \centering
    \caption{
    Closed-loop success on the RoboTwin \texttt{turn\_switch} task.
    Interaction denotes RGB--depth information flow.
    Each checkpoint is evaluated over 20 trials; overall success
    aggregates the three checkpoints.
    }
    \label{tab:depth_coevolving}
    \small
    \setlength{\tabcolsep}{4pt}
    \begin{tabular}{lccccc}
        \toprule
        \textbf{Configuration} &
        \textbf{Interaction} &
        \textbf{6K} &
        \textbf{7K} &
        \textbf{8K} &
        \textbf{Overall} \\
        \midrule
        RGB-only
        & None
        & 12/20
        & 12/20
        & 12/20
        & 60.0\% \\

        RGB + Depth
        & One-way
        & 12/20
        & 12/20
        & 14/20
        & 63.3\% \\

        \textbf{Thinker-WM}
        & \textbf{Bidirectional}
        & \textbf{15/20}
        & \textbf{15/20}
        & \textbf{15/20}
        & \textbf{75.0\%} \\
        \bottomrule
    \end{tabular}
\end{table}

As shown in Table~\ref{tab:depth_coevolving}, adding depth with one-way
interaction increases overall success from $60.0\%$ to $63.3\%$.
Bidirectional interaction achieves $75.0\%$, improving over the RGB-only
and one-way variants by $15.0$ and $11.7$ percentage points,
respectively. The full model also outperforms both variants at every
evaluated checkpoint. These results suggest that the benefit of depth
depends not only on its prediction objective, but also on how its
latent representation interacts with RGB during joint generation.

Together, these experiments demonstrate improved depth generation
accuracy on the evaluated AgiBot subset and a closed-loop control
benefit from bidirectional geometric interaction on RoboTwin.
They support treating depth as a co-evolving generative state within
the representation used for action generation.

\subsubsection{Gradient Compatibility across Function Modes}
\label{sec:experiment-interactions}

The function modes of Thinker-WM are optimized with a shared set of UDT
parameters. This design raises a basic optimization question: do the different
modes provide compatible training signals, or do some objectives interfere with
one another? We investigate this question directly through the geometry of the
loss gradients on the shared UDT.

For each function mode, we evaluate its loss on the same training batches and
with the same noise realization, while forcing the modes to run independently
on the shared weights. We compute pairwise cosine similarity between the
resulting gradients with respect to the shared UDT parameters, together with
the corresponding gradient norms. The measurement is repeated at five
checkpoints spanning the joint-training stage, so that the observed geometry
can be separated from checkpoint-specific variation.


\begin{figure}[!ht]
    \centering
    \includegraphics[width=\textwidth]{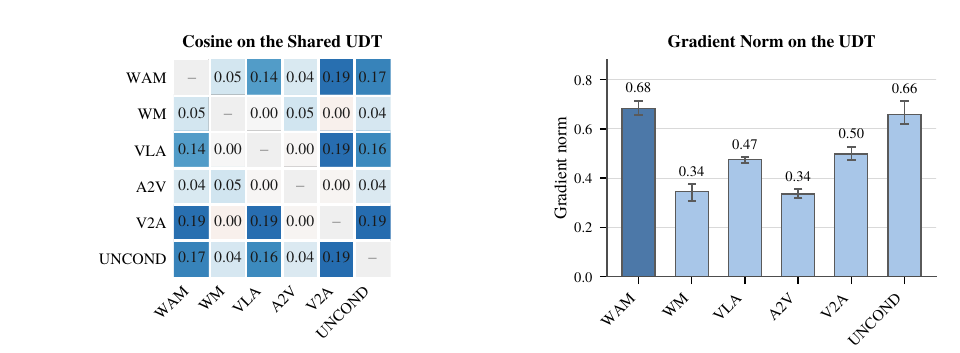}
    \caption{\textbf{Gradient geometry of the six training modes.} Every mode
is forced on its own pass over the same batches with a shared noise
realization, and the pairwise cosine of the loss gradients with respect to the
shared parameters is averaged over five checkpoints spanning the joint
training stage. Left, the cosine matrix on the shared UDT, with the diagonal
omitted. Right, the gradient norm per mode on the shared UDT, where the bar is
the mean over the five checkpoints and the whisker covers their range. No
pair is negative, and every auxiliary mode carries gradients of comparable
magnitude in a direction that adds to the joint one.}
    \label{fig:grad-cosine}
\end{figure}
Figure~\ref{fig:grad-cosine} summarizes the results. First, the shared UDT
shows no measurable pairwise gradient conflict: the minimum cosine similarity
across the six modes is approximately $-0.005$, within the variation observed
for pairs whose true alignment is near zero. Most mode pairs are therefore
either aligned or approximately orthogonal rather than negatively aligned.
Second, the degree of alignment follows the functional relationship between
the modes. The joint WAM mode shows its strongest alignment with the
inverse-dynamics, unconditional, and policy modes, with cosine similarities of
$0.19$, $0.17$, and $0.14$, respectively. In contrast, mode pairs with less
overlap in their supervised outputs remain close to zero, including $0.000$
between the WM and VLA modes and $-0.001$ between the A2V and V2A modes. This
structure is consistent with the functional overlap among the different
training objectives.

The auxiliary modes also produce substantial optimization signals on the shared
UDT. Their mean gradient norms range from $0.34$ to $0.50$, compared with
$0.68$ for the joint WAM mode. Thus, the near-orthogonal pairs do not result
from vanishing gradients; the corresponding objectives actively update the
shared representation. The same pattern is more pronounced on the action
branch, where the gradient alignment with the inverse-dynamics, unconditional,
and policy modes reaches $0.36$, $0.35$, and $0.28$, respectively. This
localization is consistent with the fact that these objectives directly
supervise the action pathway.

The action-conditioned mode provides a complementary case. Its gradient
alignment is close to orthogonal on both the shared UDT and the action branch,
with cosine similarities of $0.037$ and $0.006$, respectively. Attention
measurements on the same checkpoints show a strongly asymmetric interaction
between video and action tokens: video tokens allocate approximately $0.5\%$
of their attention mass to action tokens, whereas action tokens allocate
substantially more to video tokens. This asymmetric coupling is consistent
with the relatively weak gradient interaction of the action-conditioned
objective with the shared optimization direction.

We further verify the gradient measurements at the objective level. Under
identical batches and noise realizations, the video loss of the joint WAM mode
is $0.230$, compared with $0.232$ for the video-only WM mode, indicating that
jointly modeling action does not introduce a measurable degradation in the
video objective. Conversely, providing a ground-truth action sequence as a
clean condition reduces the video loss by $0.016$ relative to the video-only
mode, while the inverse-dynamics mode reduces the action loss by $0.010$
relative to the policy mode. These changes follow the intended role of the
conditional modes and provide an objective-level counterpart to the measured
gradient geometry.

Overall, the six modes exhibit non-conflicting and non-trivial optimization
signals on the shared UDT, while their alignment structure reflects the
functional relationships among the corresponding objectives. These measurements
support the compatibility of the multifunction training scheme with a shared
multimodal representation. They do not, however, establish whether the
particular mode sampling ratio is optimal under a fixed optimization budget,
since the six modes divide the available training steps; isolating that effect
would require matched training runs with different step allocations.

\subsubsection{Motion Probe Evaluation}
\label{sec:probe-eval}

Motion probing evaluates whether action-relevant motion information is
recoverable from the shared video representation. This analysis complements
the gradient-compatibility analysis in Section~\ref{sec:experiment-interactions}:
while the latter examines whether auxiliary training objectives interfere
during optimization, the motion probe assesses the action-related information
available in the learned representation.

We evaluate the probe on a held-out set of 96 AgiBot windows. For each UDT
checkpoint, we freeze the video representation and train the same probing-head
architecture under a consistent training protocol. The probe predicts
executed end-effector displacement from the frozen video representation alone,
and its accuracy is measured by action mean squared error (MSE). By keeping
the representation frozen and the probing protocol fixed, this evaluation
measures how readily action-relevant motion information can be extracted from
the learned features, rather than how well the representation adapts to the
probe objective.

\begin{figure}[htbp]
    \centering
    \includegraphics[width=0.5\linewidth]{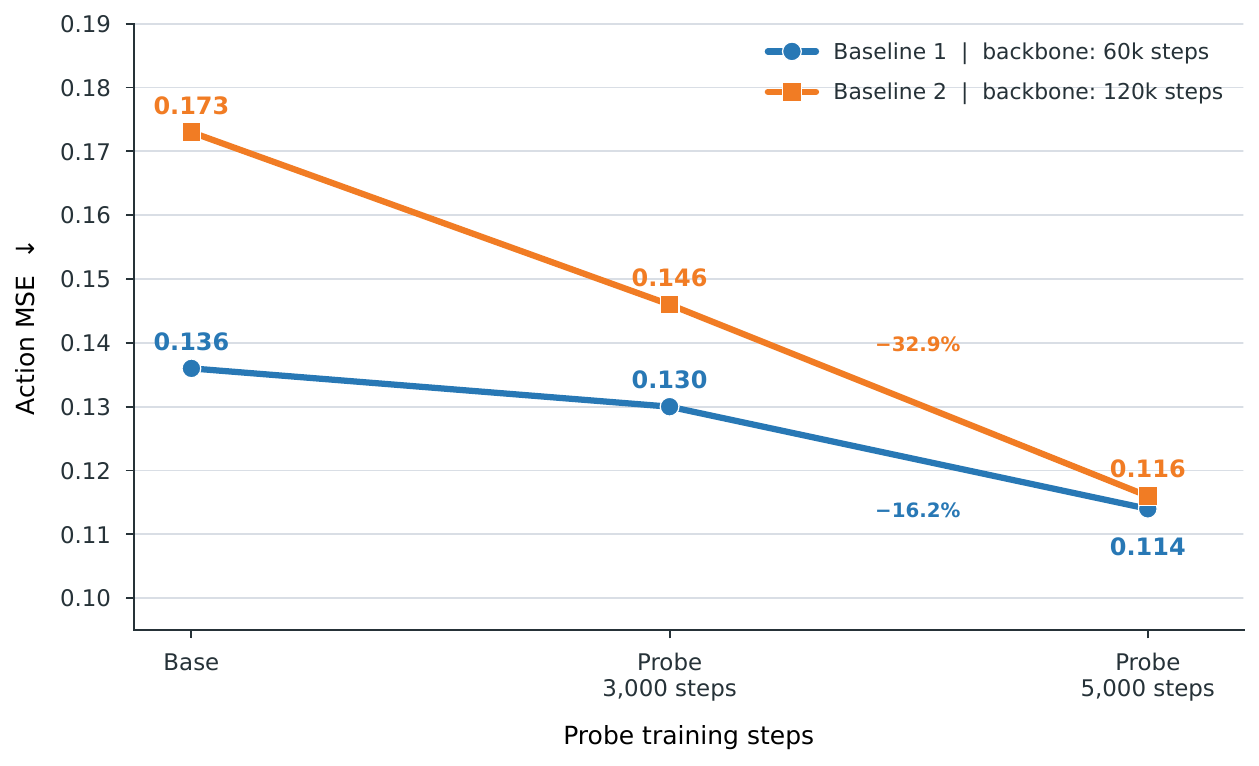}
    \caption{Action MSE of a motion probe evaluated on frozen UDT video
    representations from different training checkpoints. Lower is better.
    Both plotted series exhibit a monotonic decrease in MSE, indicating that
    action-relevant motion information becomes increasingly recoverable from
    the learned representation.}
    \label{fig:action_mse}
\end{figure}

As shown in Figure~\ref{fig:action_mse}, action MSE decreases monotonically
across the evaluated training checkpoints. This trend indicates that
end-effector displacement becomes increasingly recoverable from the frozen UDT
video representation as training progresses. The result provides evidence
that the shared representation progressively develops features that are
informative about robot motion.

The reliability of the offline visual motion teacher is important for
interpreting these results, since the probe relies on teacher-generated motion
pseudo-labels rather than manual annotations for its primary supervision.
We therefore evaluate the teacher on an in-house test split for which
ground-truth end-effector motion is available. Table~\ref{tab:probe-teacher-quality}
reports the agreement between teacher-estimated and ground-truth motion, along
with the retention rates associated with the reliability filters used to
construct the pseudo-labeled dataset.

\begin{table}[ht]
\centering
\caption{
Reliability of the offline visual motion teacher on the in-house
\texttt{test} split. The upper block reports agreement between
teacher-estimated and ground-truth end-effector motion. The lower block
summarizes the pseudo-label filtering statistics. All retention and rejection
percentages are computed relative to the full set of candidate windows.
}
\label{tab:probe-teacher-quality}
\small
\begin{tabular}{lc}
\toprule
\textbf{Metric} & \textbf{Result} \\
\midrule
\multicolumn{2}{c}{\textit{Teacher Quality}} \\
\midrule
Frame-wise directional cosine similarity
    & 0.65 \\
Cumulative displacement cosine similarity
    & 0.83 \\
End-point error (m)
    & 0.041 \\
Trajectory-shape score
    & 0.86 \\
Path correlation
    & 0.99 \\
Motion-magnitude correlation
    & 0.71 \\
\midrule
\multicolumn{2}{c}{\textit{Filtering Statistics}} \\
\midrule
Retention rate (keep)
    & 70\% \\
Retention rate (HQ)
    & 50\% \\
Rejection rate
    & 30\% \\
\bottomrule
\end{tabular}
\end{table}

The teacher achieves stronger agreement on global trajectory characteristics
than on frame-wise motion. Cumulative displacement cosine similarity reaches
0.83, while the trajectory-shape score and path correlation are 0.86 and 0.99,
respectively. The end-point error is 0.041~m. In comparison, frame-wise
directional cosine similarity is 0.65, and motion-magnitude correlation is
0.71, suggesting that local direction and displacement magnitude remain
imperfectly estimated. These results indicate that the teacher captures
important aspects of the overall motion trajectory, while leaving room for
errors at finer temporal and magnitude scales.

The keep criterion retains 70\% of candidate windows, while the stricter
high-quality (HQ) criterion retains 50\%; the remaining 30\% are rejected by
the keep filter. These statistics characterize the coverage of the filtering
procedure and the proportion of windows retained for pseudo-labeling.

Taken together, the motion-probe results show that action-relevant motion
information is recoverable from the frozen UDT video representation and becomes
increasingly accessible as training progresses. Because the probe depends on an imperfect visual teacher, these findings should be interpreted as evidence of
action-relevant information recoverability under the probe protocol, rather
than as proof of exact motion reconstruction across all candidate windows.

\subsection{Qualitative Capabilities and Generalization}

We next examine qualitative behaviors that are not fully captured by the
quantitative benchmarks above: long-horizon generation, bidirectional
interaction between trajectory and video, and zero-shot execution of unseen
instructions in unseen scenes. Rather than providing quantitative comparisons,
these experiments serve as controlled probes of whether the capabilities
suggested by the shared representation manifest at inference time.

\subsubsection{Training-Free Long-Horizon Generation}
\label{sec:experiment-autoregressive}

\begin{figure}[ht]
  \centering
  \includegraphics[width=\textwidth]{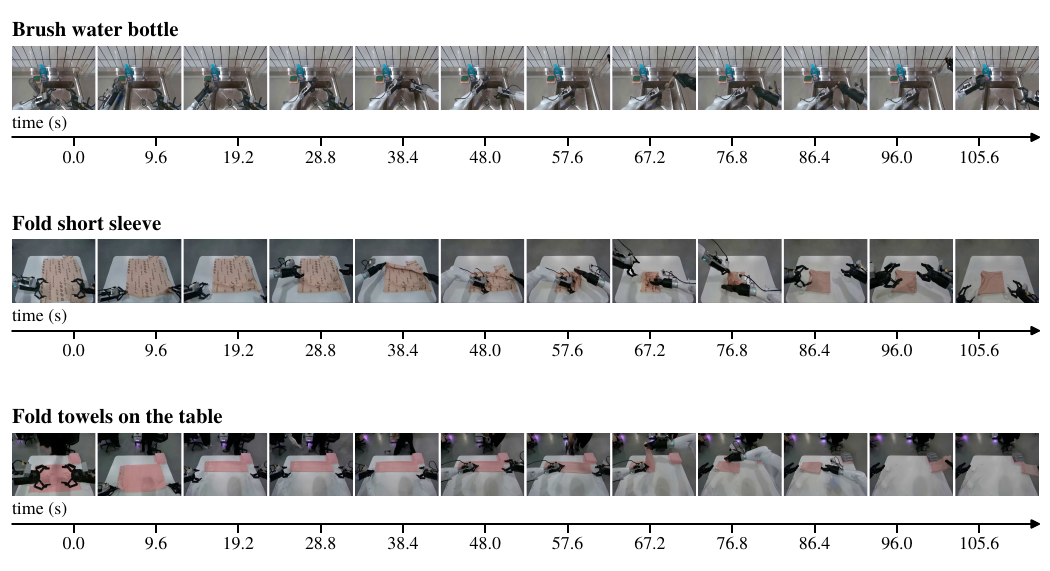}
  \caption{Long-horizon generation on three recorded episodes.
  Each block samples the frames of the model's rollout of one episode. Only the
  first frame of a rollout is a ground-truth observation; every later conditioning
  frame is produced by the model, so the $115.2$ seconds shown are generated
  without any test-time supervision.}
  \label{fig:experiment-autoregressive}
\end{figure}

The model is never trained for long rollouts. Every training window is an
independent clip of 48 frames, no stage feeds a prediction back as a condition,
and no autoregressive fine-tuning is performed. Long videos are obtained at
inference by using the last frame of one segment as the first frame of the
next, and by switching the instruction at segment boundaries so that each
segment is conditioned on the annotation covering its own time window.

We evaluate this procedure in two regimes. The first chains ten 48-frame
segments from eight EWMBench first frames with their official instructions,
yielding 96 seconds of video from a single conditioning frame. The second
chains twelve segments from sixteen long real-robot episodes, about 115 seconds
of video, with the instruction changing at every segment and each segment
covering one annotated sub-task. In both cases, the first frame is the only
ground-truth frame the model ever sees; every later conditioning frame is its
own output.

Figure~\ref{fig:experiment-autoregressive} shows three rollouts. The generation
covers several sub-tasks of a full manipulation episode, and the object layout
and robot configuration are preserved rather than drifting toward a generic
scene. The model therefore holds the scene it started from for the full rollout
without test-time supervision.

This stability is consistent with the conditioning design. The structured
instruction enters every layer through cross-attention, so each segment remains
anchored to the same task instead of recovering it from a drifting visual
context. Segment length and frame rate also match training, so continuing from
a generated frame introduces no temporal shift, and each new segment is
conditioned on a clean frame rather than a noisy latent.

\subsubsection{Trajectory Transfer and Action Recovery}
\label{sec:experiment-trajectory-transfer}



Section~\ref{sec:experiment-joint-video-action} shows that the model consumes
an action sequence when one is available. The two cases below illustrate this
behavior on recorded trajectories rather than synthetic actions.

In the first case, generation is conditioned on a real recorded trajectory. The
source is a bimanual episode dominated by left-arm motion, with the window
chosen for maximal end-effector path length. The model receives the first frame
of an evaluation sample together with that recorded trajectory, so the scene
and the trajectory come from different recordings. This tests whether the model
follows the supplied trajectory rather than reproducing the motion originally
performed in that scene.

\begin{figure}[!ht]
  \centering
  \includegraphics[width=\textwidth]{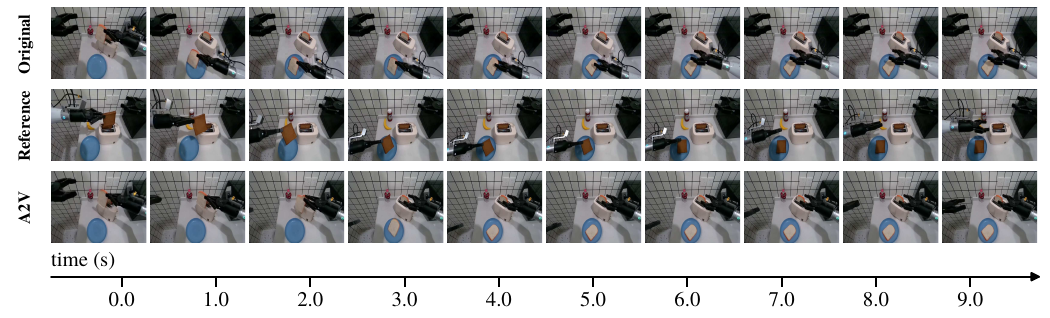}
  \caption{Trajectory transfer. Top to bottom: the original execution of the
  scene, the recorded window whose end-effector trajectory is supplied as the
  condition, and the video generated from that trajectory. Frames are sampled at
  the times marked below.}
  \label{fig:experiment-trajectory-transfer-a2v-video}
\end{figure}

\begin{figure}[!ht]
  \centering
  \includegraphics[width=\textwidth]{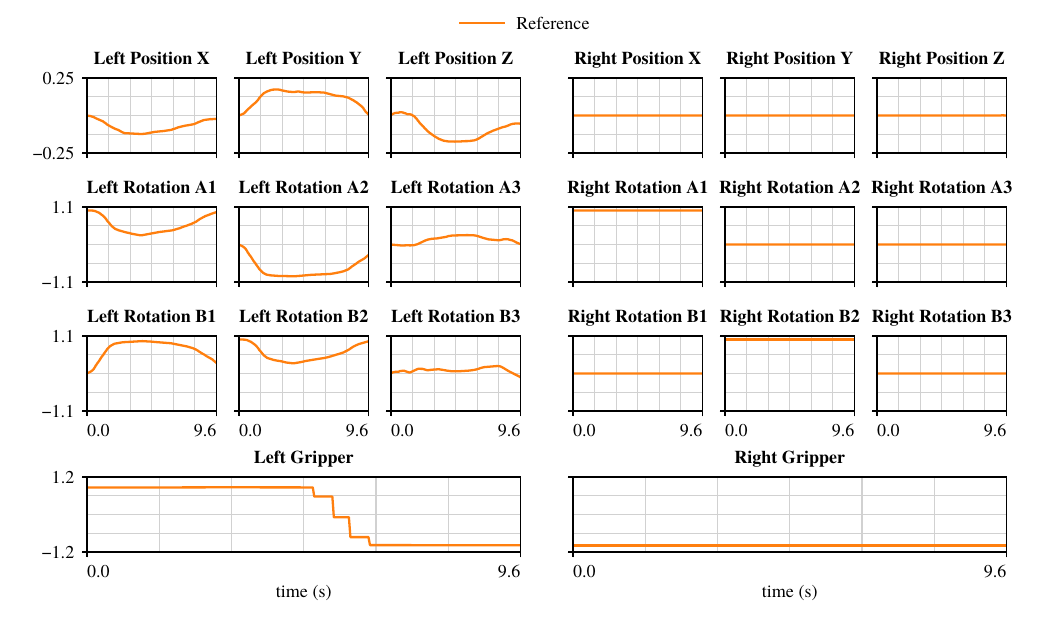}
  \caption{The end-effector trajectory supplied as the condition, in the unified
  action space. The position panels are drawn at $\pm 0.25$ and the gripper panels
  at $\pm 1.2$ so that the motion is visible; the rotation panels keep the range
  of the training monitor, $\pm 1.1$.}
  \label{fig:experiment-trajectory-transfer-a2v-action}
\end{figure}

Figure~\ref{fig:experiment-trajectory-transfer-a2v-video} compares the
generated video against both the original execution and the recording whose
trajectory was supplied. Figure~\ref{fig:experiment-trajectory-transfer-a2v-action}
plots the supplied trajectory, confirming that the generated motion follows
the requested path rather than a plausible alternative for the same task.

The second case closes the loop. The inverse mode of the same model recovers
actions from the recording by treating the video window as a clean condition
and denoising the action stream. The recovered sequence is then used to
regenerate the video. As Table~\ref{tab:experiment-v2a-rmse} reports, the
recovered action matches the recording with RMSE of $0.0540$ in native reader
space and $0.0293$ in unified action space, with the largest residuals at
gripper open/close transitions.


\begin{table}[t]
  \centering
  \caption{Accuracy of the action sequence recovered from a recorded window by
  the inverse mode, measured against the recorded trajectory.}
  \label{tab:experiment-v2a-rmse}
  \small
  \begin{tabular}{@{}lc@{}}
    \toprule
    \textbf{Space} & \textbf{RMSE}\\
    \midrule
    Native reader space, 14 dimensions & 0.0540\\
    Unified action space, 20 dimensions & 0.0293\\
    \bottomrule
  \end{tabular}
\end{table}

Together, the two cases show the same interface operating in both directions:
a recorded trajectory can drive future generation, and a recorded future can
be turned back into the trajectory that produced it. The model is therefore
not restricted to describing motion that it also invents.

\begin{figure}[!ht]
  \centering
  \includegraphics[width=\textwidth]{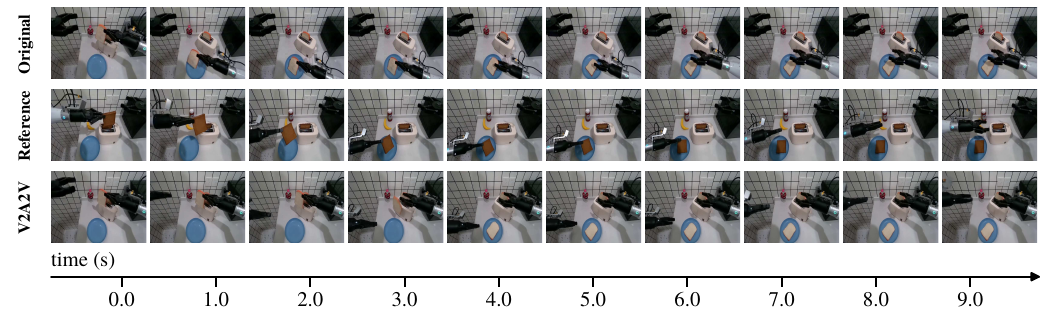}
  \caption{Closed loop over a recorded window. Top to bottom: the original
  execution of the scene, the recorded window from which the action is recovered,
  and the video generated from the recovered action. The layout follows
  Figure~\ref{fig:experiment-trajectory-transfer-a2v-video}.}
  \label{fig:experiment-trajectory-transfer-v2a2v-video}
\end{figure}

\begin{figure}[!ht]
  \centering
  \includegraphics[width=\textwidth]{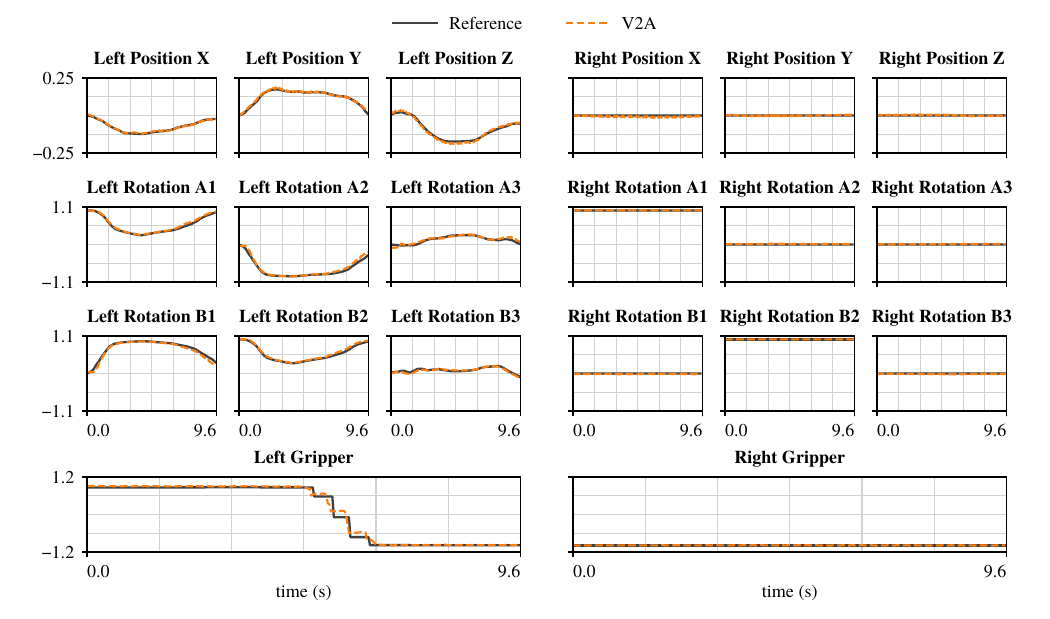}
  \caption{The action recovered from the recording by the inverse mode
  (\textsc{V2A}, dashed) against the recorded trajectory (\textsc{Reference},
  solid) over the same window, in the unified action space.}
  \label{fig:experiment-trajectory-transfer-v2a2v-action}
\end{figure}

\subsubsection{Zero-Shot Generalization}
\label{sec:experiment-zero-shot}

The cases above are drawn from the data distribution. We now remove all such
support: the conditioning frame is a hand-held photograph of a desk the model
has never seen, the instruction is written by hand rather than taken from an
annotation, and no recorded video of the scene exists, so no ground-truth
continuation is available.

\begin{figure}[!ht]
  \centering
  \includegraphics[width=\textwidth]{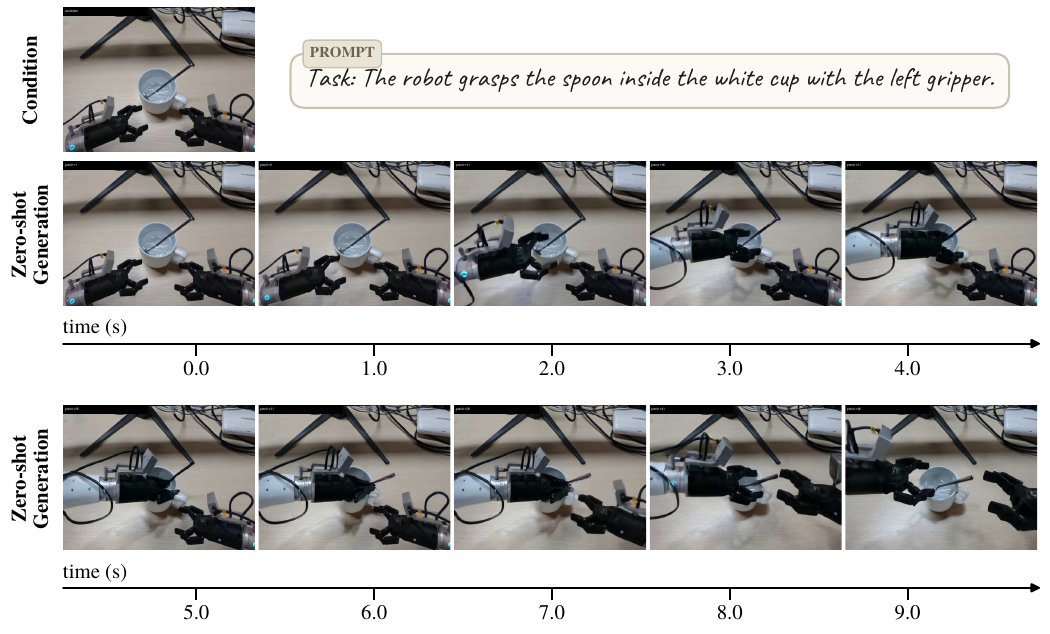}
  \caption{Zero-shot generation from a hand-held frame that is outside the
  training distribution. The prompt card repeats the instruction that was
  supplied, of which the model only sees the \texttt{Task} field.}
  \label{fig:experiment-zero-shot-a}
\end{figure}

\begin{figure}[!ht]
  \centering
  \includegraphics[width=\textwidth]{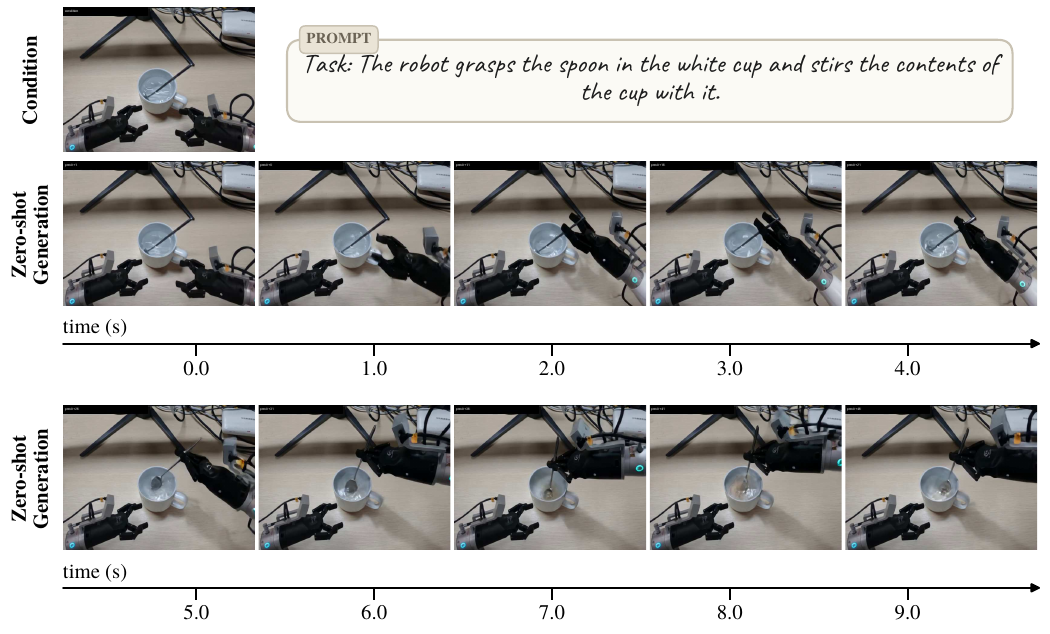}
  \caption{Zero-shot generation from the same conditioning frame as
  Figure~\ref{fig:experiment-zero-shot-a}, under the instruction given in the
  prompt card.}
  \label{fig:experiment-zero-shot-b}
\end{figure}

The two cases share the same conditioning frame and differ only in the
instruction. In the first, the left gripper must grasp a spoon handle inside a
cup and lift it out while the right arm stays still. In the second, the right
gripper must enter the same cup and stir its contents. The generated frames
carry out the requested motion in both cases
(Figures~\ref{fig:experiment-zero-shot-a} and
Figure~\ref{fig:experiment-zero-shot-b}), with the uncommanded arm remaining
in place. Since the conditioning frame is identical, the difference between
the two results must come from the instruction.

These cases isolate the two conditioning components the model was trained to
respect. The scene is unseen, so no memorized visual prior applies, and the
instruction is new, so behavior must be composed from the structured prompt
rather than retrieved. The requested motion under both conditions indicates
that the conditioning pathway generalizes beyond the annotated corpora.

\subsubsection{Zero-Shot Manipulation across Diverse Scenes}
\label{sec:Manipulation-zero-shot}

Our model is trained exclusively on demonstrations collected under
the \texttt{clean} configuration of RoboTwin, with a fixed camera
configuration and no viewpoint variation in the training data.
The baseline uses this original head-camera configuration,
with a downward viewing angle of approximately $52^{\circ}$
relative to the horizontal plane.
At evaluation time, we directly deploy the trained model in scenes
with randomized backgrounds and distractor objects, using either
the baseline view or a steeper $65^{\circ}$ view absent from training,
without additional fine-tuning.
Here, zero-shot refers to transfer to scene and camera configurations
not encountered during training, rather than to unseen task categories.

\begin{figure}[!ht]
  \centering
  \includegraphics[width=\textwidth]{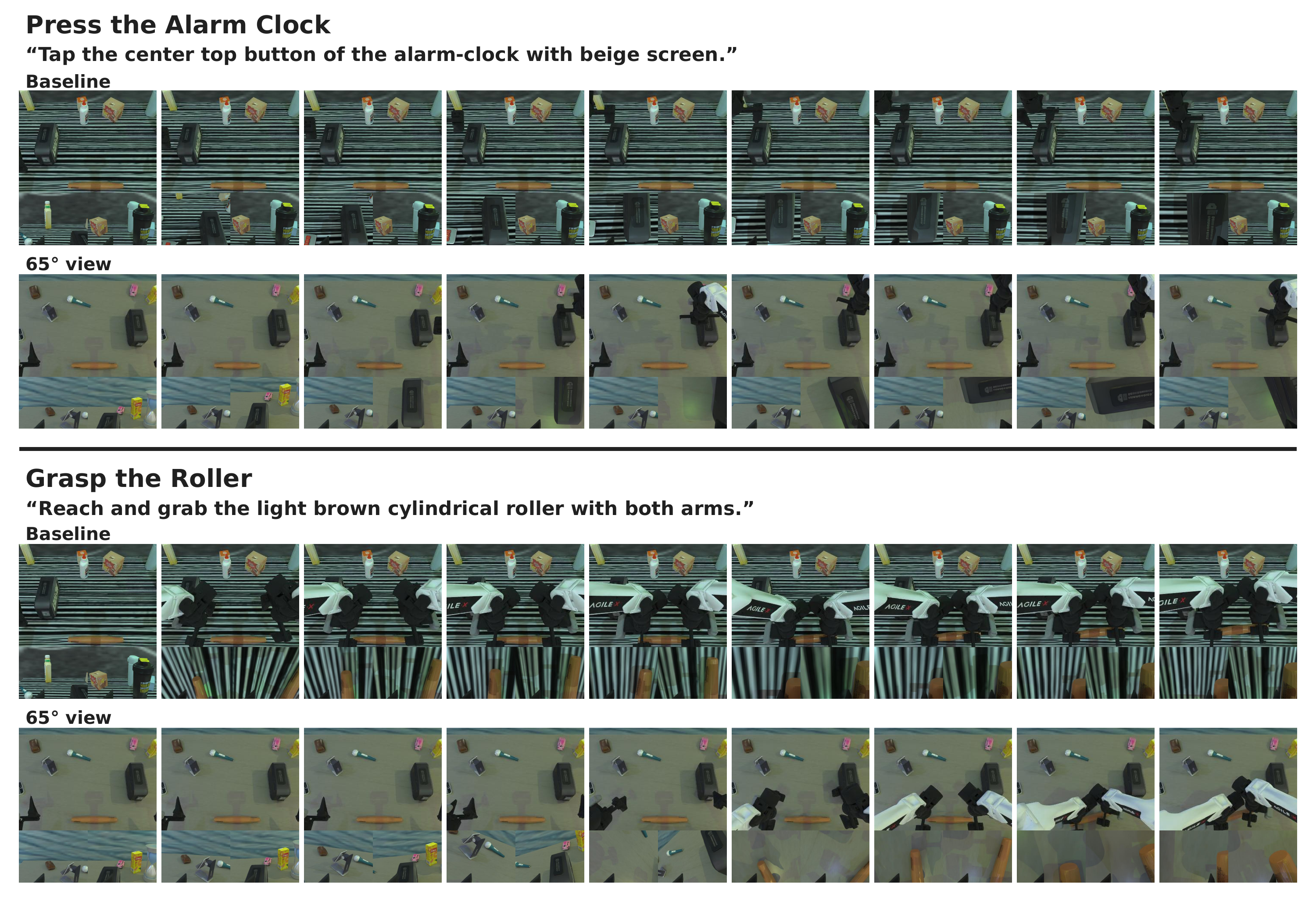}
  \caption{
      Zero-shot manipulation across diverse scenes.
  Trained only on RoboTwin \texttt{clean} demonstrations without
  viewpoint variation, the model follows new instructions to press
  an alarm clock (top) and grasp a roller (bottom) under different
  backgrounds and a $65^{\circ}$ camera view, without fine-tuning.
  Each row shows nine keyframes combining head and wrist camera views.
  }
  \label{fig:RoboTwin_Tasks_v3}
\end{figure}

Figure~\ref{fig:RoboTwin_Tasks_v3} presents successful executions of
two tasks: pressing an alarm clock and grasping a roller with both arms,
each specified using a new instruction phrasing.
The two baseline rollouts share the same initial scene and object
arrangement, allowing us to demonstrate different task executions
conditioned on the instruction while holding the visual input fixed.
The model selects and manipulates the instructed target despite
the presence of other objects. Both tasks are also successfully executed under the $65^{\circ}$
camera view in another scene with a different background.
These examples demonstrate qualitative transfer from clean,
fixed-view training data to more visually varied evaluation scenes,
including a camera configuration absent from training.
Because the background and camera configuration change together
in the displayed examples, the figure illustrates joint scene
and viewpoint generalization rather than an isolated viewpoint effect.

\begin{table}[ht]
    \centering
    \caption{Manipulation success rates (\%) on
    RoboTwin 2.0 across 50 tasks.}
    \label{tab:robotwin_success}
    \small
    \begin{tabular}{@{}lccc@{}}
        \toprule
        \textbf{Method} &
        \textbf{Clean} $\uparrow$ &
        \textbf{Randomized} $\uparrow$ &
        \textbf{Drop} $\downarrow$ 
        \\
        \midrule
        Thinker-WM (Ours) &
        \textbf{92.6} &
        \textbf{91.4} &
        \textbf{1.2} \\
        \bottomrule
    \end{tabular}
\end{table}

Beyond these qualitative examples, we further evaluate our model on all 50 tasks of RoboTwin 2.0 under the standard \texttt{Clean} and \texttt{Randomized} settings. Following the benchmark's closed-loop evaluation protocol,
we report the average task success rate across all tasks. Our model achieves \textbf{92.6\%} and \textbf{91.4\%}
success rates on \texttt{Clean} and \texttt{Randomized}, respectively, with only a 1.2 percentage-point drop, demonstrating strong robustness to visual variations.


\vspace{2cm}

\section{Conclusion and Future Work}
\label{sec:conclusion}

This work presented \textbf{Thinker-WM}, a unified multimodal world model that jointly models RGB, depth, and robot action within a shared intermediate representation. Modality-specific encoders and decoders preserve the distinct structure of each modality, while a shared Unified Dynamics Transformer (UDT) models their joint evolution with a single set of weights. This design supports multiple world-action functions under one parameterization, including joint video-action generation, world modeling, policy inference, action-conditioned generation, video-to-action generation, and inverse dynamics. Trained on approximately 20K hours of egocentric video, 10K hours of robot data, and 1K hours of simulation data, about 4K hours with aligned depth, Thinker-WM uses a staged curriculum with replay to stabilize shared representations and specialize to the deployed robot. Across benchmarks, it achieves the best Overall score on EWMBench ($4.66$), ranks first on WorldOlympiad Robotics with an All score of $0.740$, and obtains the highest 12-metric EWMScore of $71.73$ on WorldArena Track-1. Supplying a recorded action sequence as a clean condition improves Overall from $4.66$ to $5.03$, with HSD increasing from $0.5638$ to $0.7369$ and nDTW from $0.7102$ to $0.8736$. Additional experiments show that bidirectional RGB-depth interaction raises closed-loop success on RoboTwin from $60.0\%$ to $75.0\%$, and that the frozen video representation contains recoverable action-relevant motion information. These results support representation-level unification as a viable foundation for embodied world-action modeling, while also highlighting the need for further validation in real-world closed-loop settings.

Several directions remain open. First, although Thinker-WM achieves promising closed-loop control in simulation (e.g., RoboTwin), its effectiveness as a closed-loop policy on real robots remains to be established. Conducting real-world closed-loop manipulation experiments, evaluating robustness to real-world perturbations, and investigating sim-to-real transfer of the shared action interface are critical next steps. Second, geometric supervision covers only a subset of the training mixture; extending depth-aware learning to larger collections of egocentric and general video could further strengthen physical representation learning. Third, long-horizon generation is currently performed through inference-time autoregressive rollout; training directly on model-generated trajectories and quantitatively evaluating long-horizon drift remain open. Finally, scaling the model and training mixture, understanding how the geometric channel, data composition, and training curriculum should evolve with scale, and optimizing the sampling ratios of the six function modes under a fixed optimization budget are additional directions for future study.

\vspace{2cm}
\noindent\section*{Contributors}
\addcontentsline{toc}{section}{Contributors}

\newcommand{\contributor}[2]{%
  \par\noindent
  \begin{minipage}{\linewidth}
    \setlength{\parindent}{0pt}%
    \setlength{\parskip}{0pt}%
    \setlength{\leftskip}{0pt}%
    \setlength{\rightskip}{0pt}%
    \noindent{\color{ubdark}\bfseries\small\MakeUppercase{#1}}\par
    \if\relax\detokenize{#2}\relax\else
      \vspace{0.15em}%
      \noindent{\small #2}\par
    \fi
  \end{minipage}%
  \par\vspace{0.45em}%
}

\contributor{Core Author}{%
{\hypersetup{hidelinks}\href{https://orcid.org/0009-0008-9791-3733}{Chen Liao}}, Jiyuan Wang, \hypersetup{hidelinks}\href{https://grandzxw.github.io}{Xiongwei Zhao}
}
\hspace{1cm}
\contributor{Author}{%
Peng Gao, \hypersetup{hidelinks}\href{https://scholar.google.com/citations?hl=zh-CN\&user=yycQdx4AAAAJ}{Yiming Ma}, Daqin Luo, Xiangming Xu
}
\hspace{1cm}
\contributor{Project Lead}{Baiyu Pan}
\hspace{1cm}
\contributor{Advisor}{Jichao Jiao, Hailin Shi}

\vspace{2cm}
\section*{Acknowledgments}
\addcontentsline{toc}{section}{Acknowledgments}

We would like to express our sincere gratitude to the Cloud Platform Team and the IT Team for their generous support, especially to Wei Sun, Junhui Huang, Yang Peng, Dong Liang, Wencai Deng for their valuable contributions.

\newpage
\bibliography{refs}

\end{document}